\documentclass[letterpaper]{article} 
\usepackage[submission]{aaai2026}  
\usepackage{times}  
\usepackage{helvet}  
\usepackage{courier}  
\usepackage[hyphens]{url}  
\usepackage{array}
\usepackage{booktabs} 
\usepackage{placeins}
\usepackage{graphicx} 
\usepackage{times}
\usepackage{tabularx}
\usepackage{helvet}
\usepackage{courier}
\usepackage{float}
\usepackage{xcolor}
\usepackage{url}
\usepackage{natbib}  
\usepackage{caption} 
\usepackage{algorithm}
\usepackage{algorithmic}

\usepackage{newfloat}
\usepackage{listings}
\usepackage{subcaption}
\usepackage{subfig}
\usepackage{xcolor}
\usepackage{amsmath}
\DeclareCaptionStyle{ruled}{labelfont=normalfont,labelsep=colon,strut=off} 
\floatstyle{ruled}
\newfloat{listing}{tb}{lst}{}
\floatname{listing}{Listing}
\title{Punctuation and Predicates in Language Models}
\author{
    Sonakshi Chauhan\textsuperscript{1} \footnote{Correspondence to: sonakshichauhan1402@gmail.com}, 
    Maheep Chaudhary\textsuperscript{2} , 
    Koby Choy\textsuperscript{2}, 
    Samuel Nellessen\textsuperscript{3}, 
    Nandi Schoots\textsuperscript{4}\\
}
\affiliations{
    \textsuperscript{\rm 1}Univesity of Glasgow,
    \textsuperscript{\rm 2}Independent,
    \textsuperscript{\rm 3}Radboud University Nijmegen,
    \textsuperscript{\rm 4}University of Oxford
}

\begin{document}

\maketitle

\begin{abstract}
In this paper we explore where information is collected and how it is propagated throughout layers in large language models (LLMs). 
We begin by examining the surprising computational importance of punctuation tokens which previous work has identified as attention sinks and memory aids. 
Using intervention-based techniques, we evaluate the necessity and sufficiency (for preserving model performance) of punctuation tokens across layers in GPT-2, DeepSeek, and Gemma. 
Our results show stark model-specific differences: for GPT-2, punctuation is both necessary and sufficient in multiple layers, while this holds far less in DeepSeek and not at all in Gemma. 
Extending beyond punctuation, we ask whether LLMs process different components of input (e.g., subjects, adjectives, punctuation, full sentences) by forming early static summaries reused across the network, or if the model remains sensitive to changes in these components across layers. 
Extending beyond punctuation, we investigate whether different reasoning rules are processed differently by LLMs.
In particular, through interchange intervention and layer-swapping experiments, we find that conditional statements (if, then), and universal quantification (for all) are processed very differently.
Our findings offer new insight into the internal mechanisms of punctuation usage and reasoning in LLMs and have implications for interpretability. \\



Code:
\url{https://github.com/SonakshiChauhan/reasoning}
\end{abstract}





\section{Introduction}


Recent work has revealed that LLMs often perform tasks in ways that diverge significantly from human reasoning. 
In particular, punctuation which is often considered negligible in human processing appears to play a surprisingly active and complex role within these models. 
Studies have shown that punctuation tokens can act as attention sinks \cite{gu2025attentionsinkemergeslanguage, barbero2025llmsattendtoken}, increase computational memory \cite{pfau2024letsthinkdotdot}, and even serve as information carriers \cite{tigges2023linearrepresentationssentimentlarge}. 
Despite their seemingly minor status, punctuation marks appear to shoulder a disproportionately large computational burden.

In this work, we examine the role of punctuation in summarization and information propagation through the lens of necessity and sufficiency: how essential are these tokens for preserving meaning and reasoning within the model?
We use zeroing out interventions to better understand how punctuation influences layerwise computation dynamics. 
To test whether the period acts as a boundary for information summarization in the context of reasoning, we also perform targeted activation interchange interventions. 

Building on this, we take a mechanistic lens to ask a broader question: How do LLMs internally represent and process reasoning? 
Previous work on reasoning has investigated reasoning rules using graph-based approaches \cite{luo2024chatrulemininglogicalrules,qu2021rnnlogiclearninglogicrules,tang2024ruleknowledgegraphreasoning}.
In this paper, we investigate whether models interpret inputs compositionally, treating different elements of a sentence (such as subjects, adjectives, punctuation, and full syntactic structures) as functionally distinct units or whether they instead form a static summary early in the forward pass and reuse that across layers. 
This line of inquiry helps us understand not just whether a model can reason, but how that reasoning unfolds internally. 
This distinction is an effort to zoom-in on how closely model behaviour aligns or diverges from human reasoning patterns.

To understand how rules are processed by language models, we analyze which layers are responsible for reasoning over different types of logical and syntactic structures. 
Specifically, we aim to identify whether rule-consistent behavior is distributed across the model or concentrated in particular layers.
We use part-of-sentence interventions and layer swaps, with the aim of characterizing how different layers contribute across syntactic and logical dimensions.

Our main findings are:
\begin{itemize}
\item 
For GPT-2 we find that period and question mark tokens are necessary and sufficient in five out of twelve layers. 
However, in DeepSeek this only holds for one in twentyfour layers, and in Gemma there is no layer for which this holds. See Section \ref{sec:punctuation-results}.
We are the first to investigate sufficiency and necessity of punctuation tokens, which we do by selectively (non-)zeroing out tokens.
Additionally, although interchange interventions have been done on tokens, they were not previously used to investigate attention sinks and memorization.

\item For the models we investigate, we find different layerwise responses to interventions on conditional statements and universal quantification, which are two types of reasoning rules. See Section \ref{sec:reasoning-results}.
We are the first to investigate conditional statements and universal quantification by applying interchange intervention.

\item We investigate which layers are most swappable, and find different LLMs behave very differently. 
For GPT-2 and Gemma we find that middle layers are unique, for DeepSeek we find that layers become more and more interchangeable, when prompted on statements containing reasoning rules. See Figure \ref{fig:all_heatmaps}.
We are the first to compare reasoning rules using layer swaps.
\end{itemize}

Across models there is a wide variation in necessity and sufficiency of punctuation, 
sensitivity to interventions on different reasoning rules, 
and non-uniform swappability of layers.
This reveals sharp differences in how internal computations are organized across architectures or training paradigms, and suggests distinct underlying inductive biases in how models learn to represent and perform reasoning. 



\section{Related Work}


 Mechanistic interpretability seeks to reverse-engineer neural networks by linking  neural components to specific behaviors and computations \cite{olah2017feature}.
 In causal abstraction this linking is done by forming a causal model of the behaviors \cite{geiger2021causalabstractionsneuralnetworks}.
 The following lines of research are most related to our work.

\paragraph{Layerwise Process Analyses.}
Previous work has shown earlier layers process initial structure of the input \cite{yang2025internalchainofthoughtempiricalevidence}, and ablating them leads to substantial degradation \cite{zhang2024investigatinglayerimportancelarge}.
Middle layers have been found to build intermediate representations \cite{yang2025internalchainofthoughtempiricalevidence}, to be essential for a variety of reasoning tasks \cite{skean2025layer, liu2025middlelayerrepresentationalignmentcrosslingual}, and they often operate in a shared representational space, swapping them has little effect for general tasks but this does not hold true for structured domains like math and logical reasoning \cite{sun2025transformerlayerspainters}. 
Later layers focus on task-specific specialization \cite{yang2025internalchainofthoughtempiricalevidence}, 
contain attention heads specialized for generalization \cite {ye2025functioninductiontaskgeneralization},
and may primarily serve to consolidate and finalize already-inferred information \cite{benartzy2024attendfirstconsolidatelater}.


\paragraph{Attention Sinks}
emerge naturally during pretraining giving disproportionate attention to the first token \cite{gu2025attentionsinkemergeslanguage}. 
Such attention sinks may be functionally necessary to prevent representational collapse and ensure smooth information flow in long-context scenarios \cite{barbero2025llmsattendtoken}. 
\citet{tigges2023linearrepresentationssentimentlarge} demonstrate that punctuation symbols act as aggregation points for sentiment signals within the model’s activation space. 
Punctuation tokens serve as structural anchors, influencing factual consistency, context retention, and reasoning \cite{razzhigaev2025llmmicroscopeuncoveringhiddenrole, zhu2025layercaketokenawarecontrastivedecoding}. 
Punctuation may also act as computation enablers, providing additional ``thinking steps" irrespective of their linguistic function \cite{pfau2024letsthinkdotdot}. 

\paragraph{Reasoning.}

Techniques like steering vectors \cite{venhoff2025understanding}, chain-of-thought prompting \cite{plaat2024reasoninglargelanguagemodels, dutta2024thinkstepbystepmechanisticunderstanding} and multi-hop reasoning analysis \cite{wang2024buffermechanismmultistepinformation} aim to make the implicit reasoning steps of LLMs more interpretable and controllable.
LLMs may construct internal reasoning trees, recoverable via attention-based probes \cite{hou2023towards}, indicating genuine multi-hop reasoning beyond memorization. 
Benchmarks like Cladder \cite{jin2024cladderassessingcausalreasoning} systematically evaluate a model's ability to extract and utilize causal structures from text. 
\citet{bogdan2025thoughtanchorsllmreasoning} perform sentence-level reasoning analysis using thought vectors, revealing that a small subset of these vectors disproportionately governs the model’s reasoning process.
\citet{ibeling2021openuniversecausalreasoning, chauhan2024gpt} have explored reasoning from a causal perspective.

\section{Models and Datasets}

\paragraph{Models:} We conduct our experiments using three models of varying sizes and architectures: GPT-2 Small \cite{lee2019patentclaimgenerationfinetuning}, Gemma \cite{gemmateam2024gemma2improvingopen}, and DeepSeek \cite{guo2024deepseekcoderlargelanguagemodel}. A clear distinction in the characteristics of the models can be seen in Table \ref{tab:model_comparison} in the appendix. 

We use these models to answer questions based on the context provided; each question can have three possible answers 1. \textit{True}, 2. \textit{False}, 3. \textit{Unknown}. 
Our main aim of having an experimental setup like this is to check how the model processes context and then check the understanding of the context with the help of questions.

\paragraph{RuleTaker Dataset} \cite{clark2020transformerssoftreasonerslanguage}  tests the reasoning and implication abilities of LLMs. 
It includes facts and rules, followed by questions that assess whether the rules are correctly applied. 
Answers to these questions are labeled as True, False, or Unknown. 

An example prompt is:
\textit{``Harry is tall. Tall people are round. Is Harry round?''}
In the above example, the first sentence is a fact, the second sentence is a (universal quantification) rule, followed by a question that the model answers. 

\paragraph{Fine-tuning on RuleTaker Dataset:} We fine-tuned GPT2, DeepSeek, and Gemma on the ruletaker dataset. 
We fine-tune all models to specialize their knowledge for our targeted reasoning task, which involves structured logical rules beyond the scope of general pre-training. 
Furthermore, since our data set is formulated as a supervised classification problem with three labels, task-specific fine-tuning is necessary to align the outputs of the models with the label space. 
While GPT-2 is fine-tuned using full parameter training, we apply LoRA \cite{hu2021loralowrankadaptationlarge} for Gemma and DeepSeek due to their larger sizes, allowing efficient adaptation with minimal parameter updates. 
After finetuning, GPT-2 and Gemma achieve 96\% classification accuracy on a held-out validation set, while DeepSeek achieves 93\%.

\paragraph{Interchange Intervention Dataset:} We curate subsets of the original RuleTaker dataset to assess model predictions and interchange intervention effectiveness. 
These datasets follow the format: 
\textit{base prompt}, \textit{override prompt}, \textit{base answer}, \textit{override answer}, \textit{question}. 
The model is prompted with `\texttt { \textless base\textgreater{} Question: \textless question \textgreater{} }'. 
The \textit{base answer} is the model's original response, while the \textit{expected answer} is what it should output after a successful interchange intervention. 
Questions are designed based on the type of interchange intervention performed, where we have questions that check the base information is removed and questions that check whether the information from the override prompt has been introduced.
Here is an idealized example (without distractor sentences):
\begin{quote}
\textbf{Base prompt:} Rabbit like squirrel. If something like squirrel then squirrel chase rabbit. \\
\textbf{Override prompt:} Anne is young. If someone is quiet then they are young.\\
\textbf{Question:} Squirrel chase rabbit? \\
\textbf{Base Answer:} \textit{True} \quad \textbf{Override Answer:} \textit{Unknown}
\end{quote}

We design separate subsets targeting Conditional Rules and Universal Quantification. 

\section{Methodology}

\subsection{Interventions}
\label{sec:interventions}

Below we describe our interventions, which we use different terms for, even when they can be seen as special cases of each other.

For each of these interventions, we intervene on the residual blocks.
In particular, we intervene on the output from one residual block component, which acts as the input to the next residual block. 
We do this because the combined effect of attention and MLP components are captured in the block output,
and we want to intervene on a ``macro'' layer behaviour as opposed to smaller subcomponents of the layer.

\emph{Zeroing-out Intervention:}
\label{sec:zeroing_out}
\cite{mcgrath2023hydraeffectemergentselfrepair}
Here we selectively replace token activations with zero. 
We perform two versions of this experiment, in one version, which we call \textit{zero} (or necessity) intervention, a token activation $z$ is zeroed out, and all other activations are left untouched.
In a second version, which we call \textit{non-zero} (or sufficiency) intervention, we leave a token activation $z$ untouched, and zero out all other activations.

\emph{Interchange Intervention}
\cite{geiger2022inducingcausalstructureinterpretable} 
involves replacing the internal activation of the \textit{base prompt} with that of the \textit{override prompt} at a specific intervention layer and token position. 
Formally, let $\mathbf{z}^{(l)}_{\text{base}, t}$ denote the output of the transformer block at layer $l$ and token position $t$ for the base prompt. We replace this with $\mathbf{z}^{(l)}_{\text{override}, t}$ the corresponding activation from the override prompt and continue the forward pass using this modified representation.

This allows us to examine how information originating from the override prompt influences the model's response when introduced partway through computation on the base prompt. By performing interchange interventions across layers $l = 0, 1, ..., L$, we identify which parts of the network are most responsible for reasoning over logical structures.

We use the NNsight library \citep{fiottokaufman2024nnsightndifdemocratizingaccess} to execute the interchange interventions.


\emph{Layer Swap Intervention:}
\cite{lad2025remarkablerobustnessllmsstages}
We take the logit with the highest value, and check if it corresponds to True, False, or Unknown.
We only considered datapoints where the model originally answers questions correctly, i.e. the logit for the correct class (True, False, Unknown) was highest.
To do a layer swap intervention, we replace the entire layer activation of the  intervention layer, with the entire activation of the swap layer. 


\subsection{Evaluation Metrics}\label{sec:metrics}


\emph{Zeroing Out Accuracy}
After zeroing out tokens as described in Section \ref{sec:zeroing_out}, we obtain a predicted label by applying argmax over the output logits. 
Accuracy for each layer is computed as a fraction of samples for which predicted label matches the ground-truth label. 

\emph{Interchange Intervention Accuracy (IIA)}
measures whether a model's response aligns with the expected outcome after an interchange intervention. 
For each input, we perform interchange intervention as defined in Section \ref{sec:interventions}. 
The model is then queried with a follow-up question, and its answer is compared to the \textit{expected answer} (i.e., the answer the model would give if it fully adopted the logic from the override prompt).

Let $y^{(i)}$ be the model's answer to the $i$-th intervened example, and let $\hat{y}^{(i)}$ be the corresponding \textit{override answer}. Then IIA is defined as
$\frac{1}{N} \sum_{i=1}^{N} \mathbf{1}_{\left[y^{(i)} = \hat{y}^{(i)}\right]}
$, where $\mathbf{1}_{[\cdot]}$ is the indicator function and $N$ is the number of examples.

This metric captures whether the interchange intervention successfully transfers logical structure from the override prompt into the base context. By computing IIA across layers, we identify where in the model such logical reasoning is most causally encoded.

\emph{Layer Swap Accuracy:} We calculate the impact of the layer swap interchange intervention by taking the difference between the logit of the correct class (True, False, Unknown) when using the original layer’s activations during inference, and the logit for the correct class when using the swapped activation during inference.

\subsection{Intervention Targets}\label{sec:targets}

\subsubsection{Punctuation Zeroing Out} targets are all the periods in a prompt, and/or the question mark in a prompt.

\paragraph{Punctuation Interchange Interventions} targets are:
\begin{itemize}
    \item \textbf{First Sentence}: For a base prompt, consider the first full sentence excluding the period. Swap the activations of that entire base sentence with those of an override.
    \item \textbf{First Period}: For a base prompt, consider the first period. Swap the activations at that period between a base sentence and an override sentence.
\end{itemize}

Let the base be \textit{``Dave is nice. Fiona is grey. If someone is happy, they are cool.''} and the override be \textit{``Ben is purple. Adam is cool. All happy are sad.''}
In the first-period swap, we exchange the activation at the period following \textit{``Dave is nice"} with that from the period after \textit{``Ben is purple"}. 
We then ask a question derived from the override context, such as \textit{“Is Ben purple?”} 
A correct response indicates that the information from the base context has been transferred via the period token — suggesting it acts as a summarization.


\begin{figure*}[h]
    \centering
    \textbf{Sufficiency (selectively non-zero)}\par\medskip
    \begin{subfigure}[b]{0.32\linewidth}
        \includegraphics[width=\linewidth]{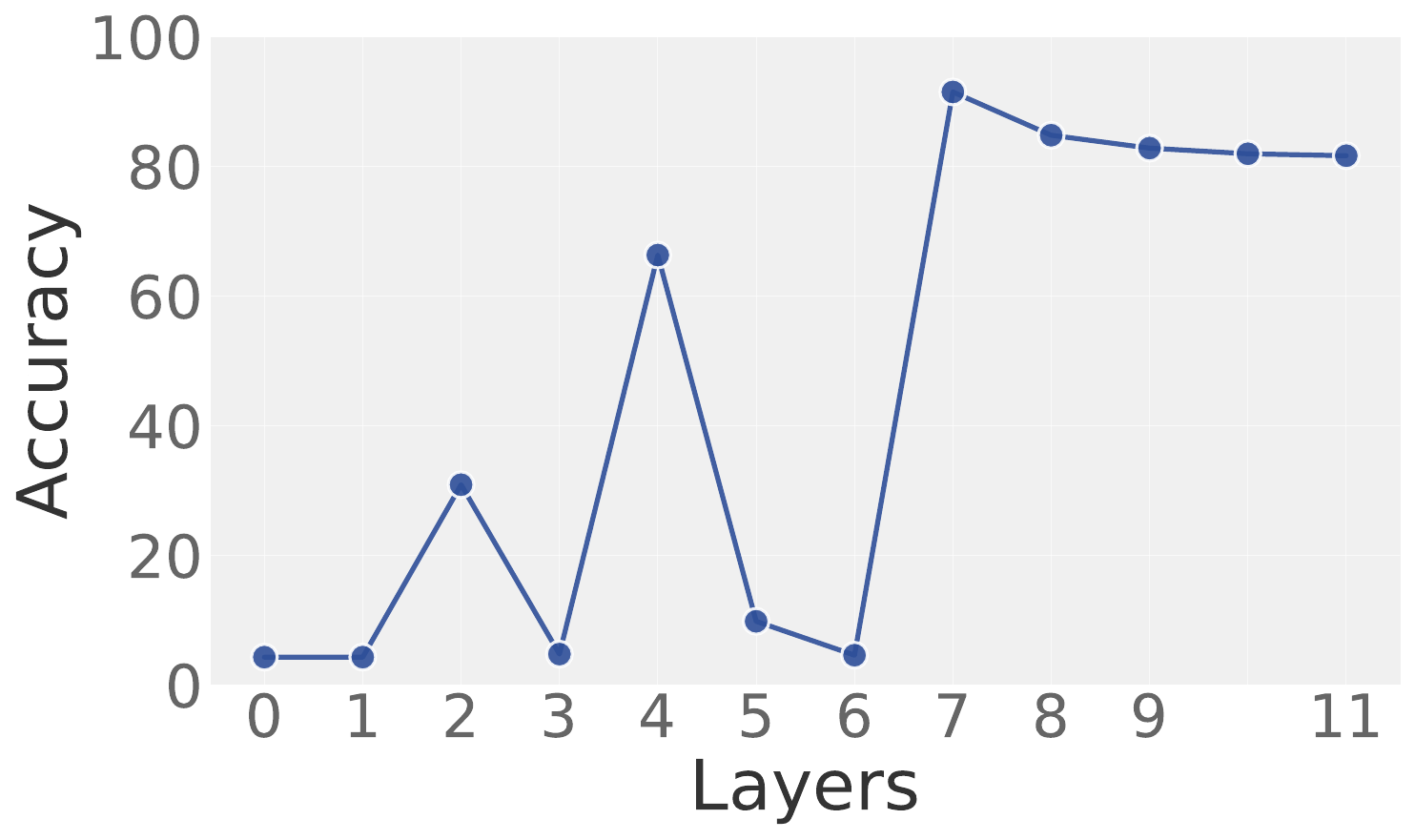}
        \caption{GPT2}
        \label{fig:non-zero-gpt2-period-and-question}
    \end{subfigure}
    \hfill
    \begin{subfigure}[b]{0.32\linewidth}
        \includegraphics[width=\linewidth]{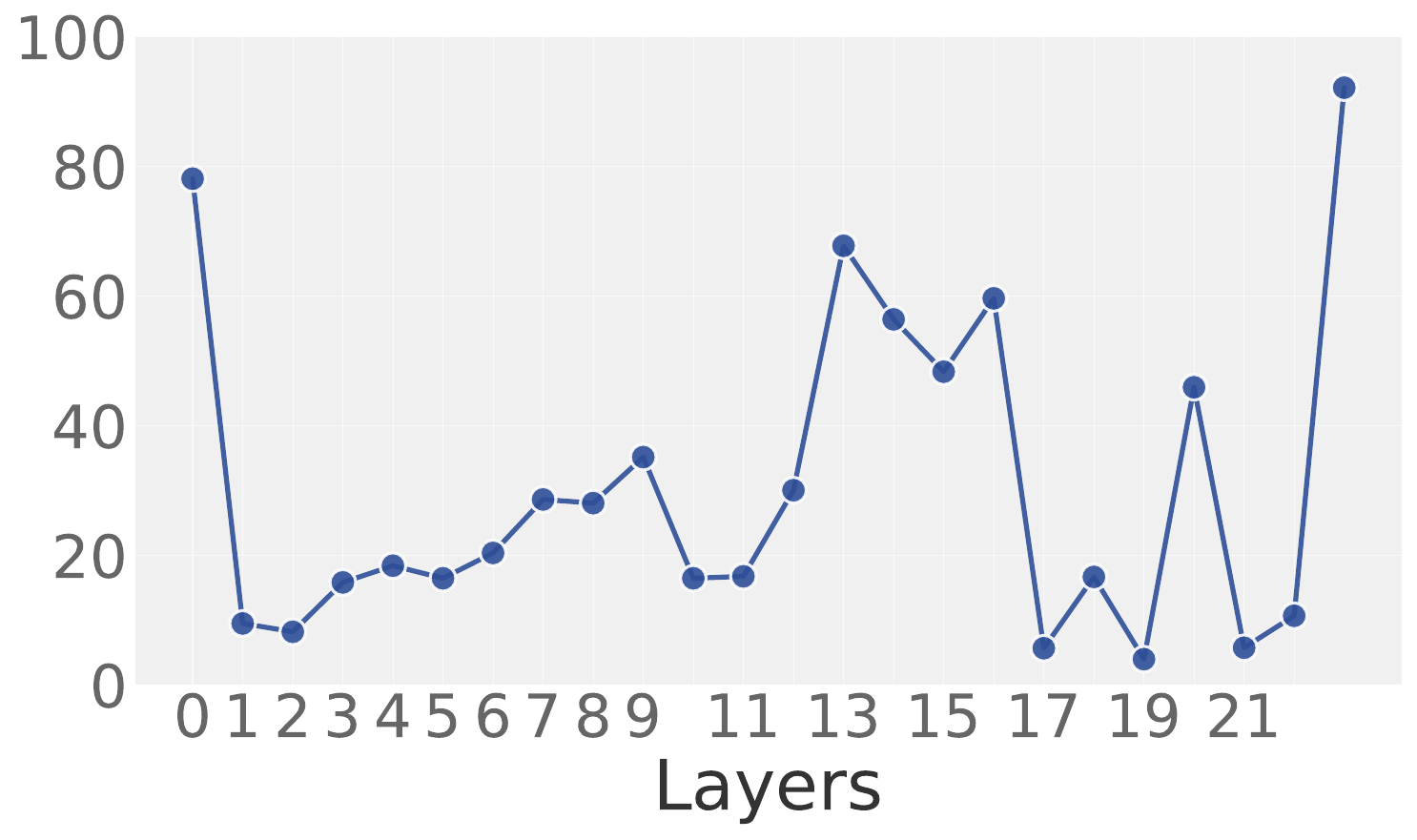}
        \caption{DeepSeek}
        \label{fig:nonzero-deepseek}
    \end{subfigure}
    \hfill
    \begin{subfigure}[b]{0.32\linewidth}
        \includegraphics[width=\linewidth]{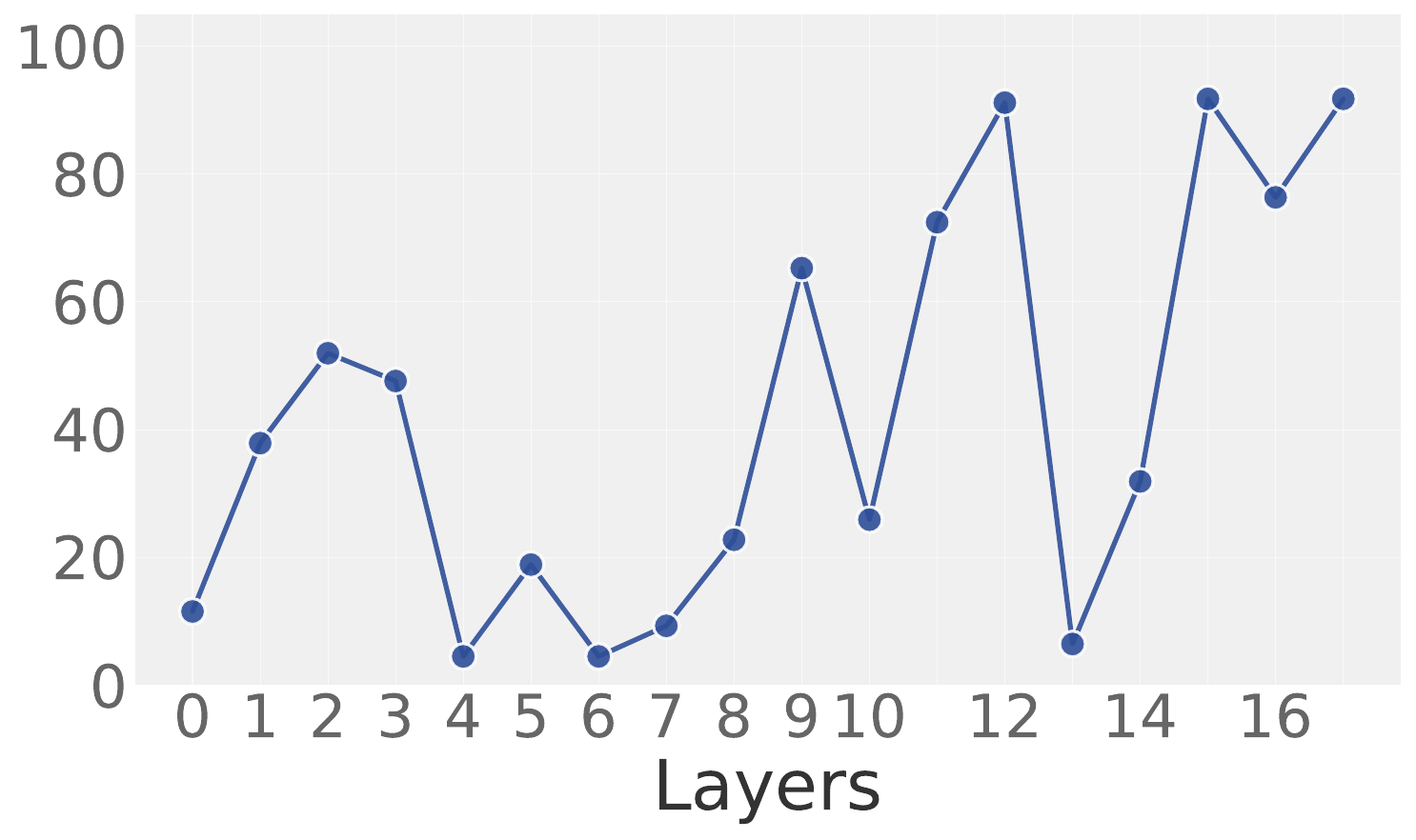}
        \caption{Gemma}
        \label{fig:nonzero-gemma}
    \end{subfigure}
    
    \vspace{0.5cm}
    \textbf{Necessity (selectively zero)}\par\medskip
    \begin{subfigure}[b]{0.32\linewidth}
        \includegraphics[width=\linewidth]{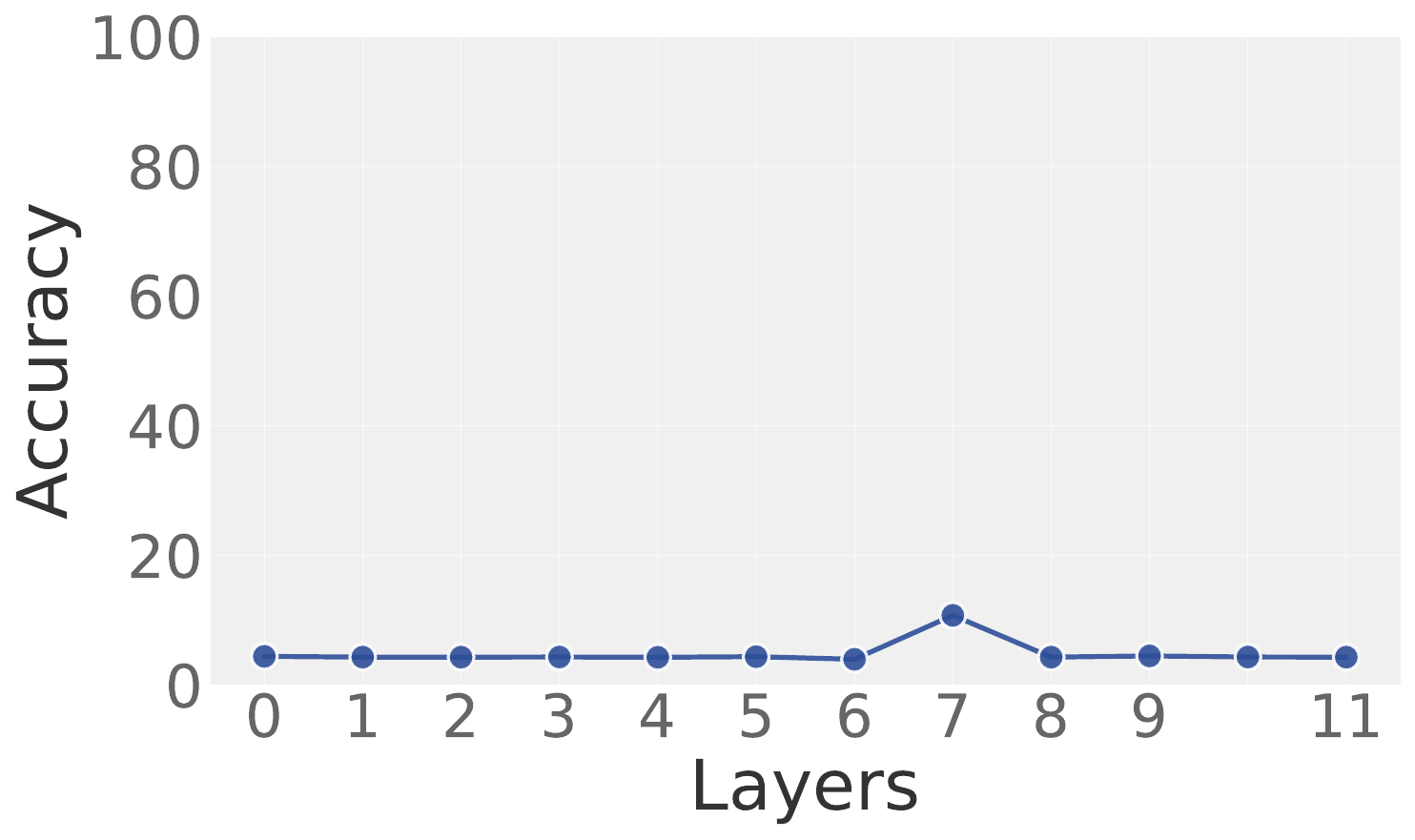}
        \caption{GPT2}
        \label{fig:zero-gpt2-period-and-question}
    \end{subfigure}
    \hfill
    \begin{subfigure}[b]{0.32\linewidth}
        \includegraphics[width=\linewidth]{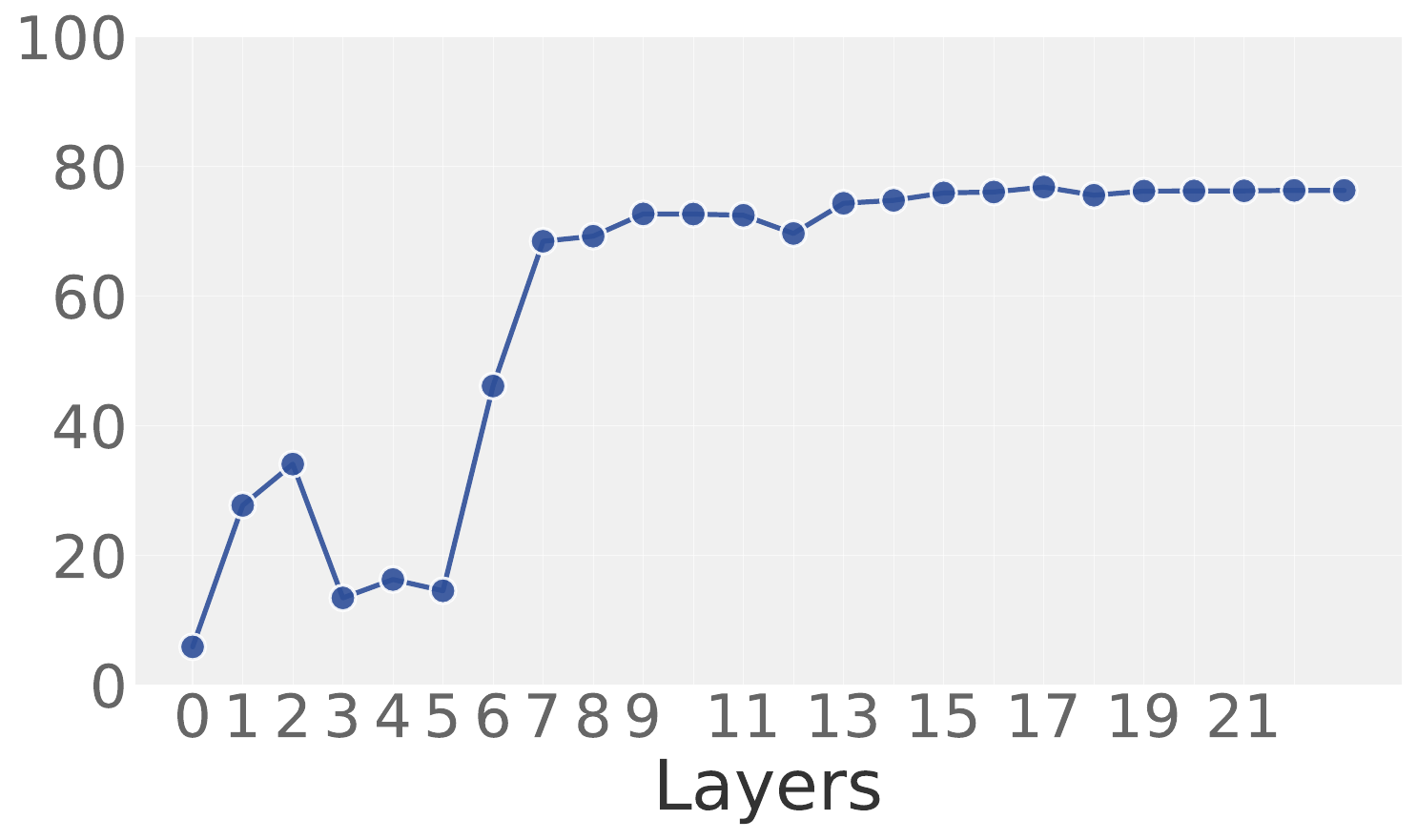}
        \caption{DeepSeek}
        \label{fig:zero-deepseek}
    \end{subfigure}
    \hfill
    \begin{subfigure}[b]{0.32\linewidth}
        \includegraphics[width=\linewidth]{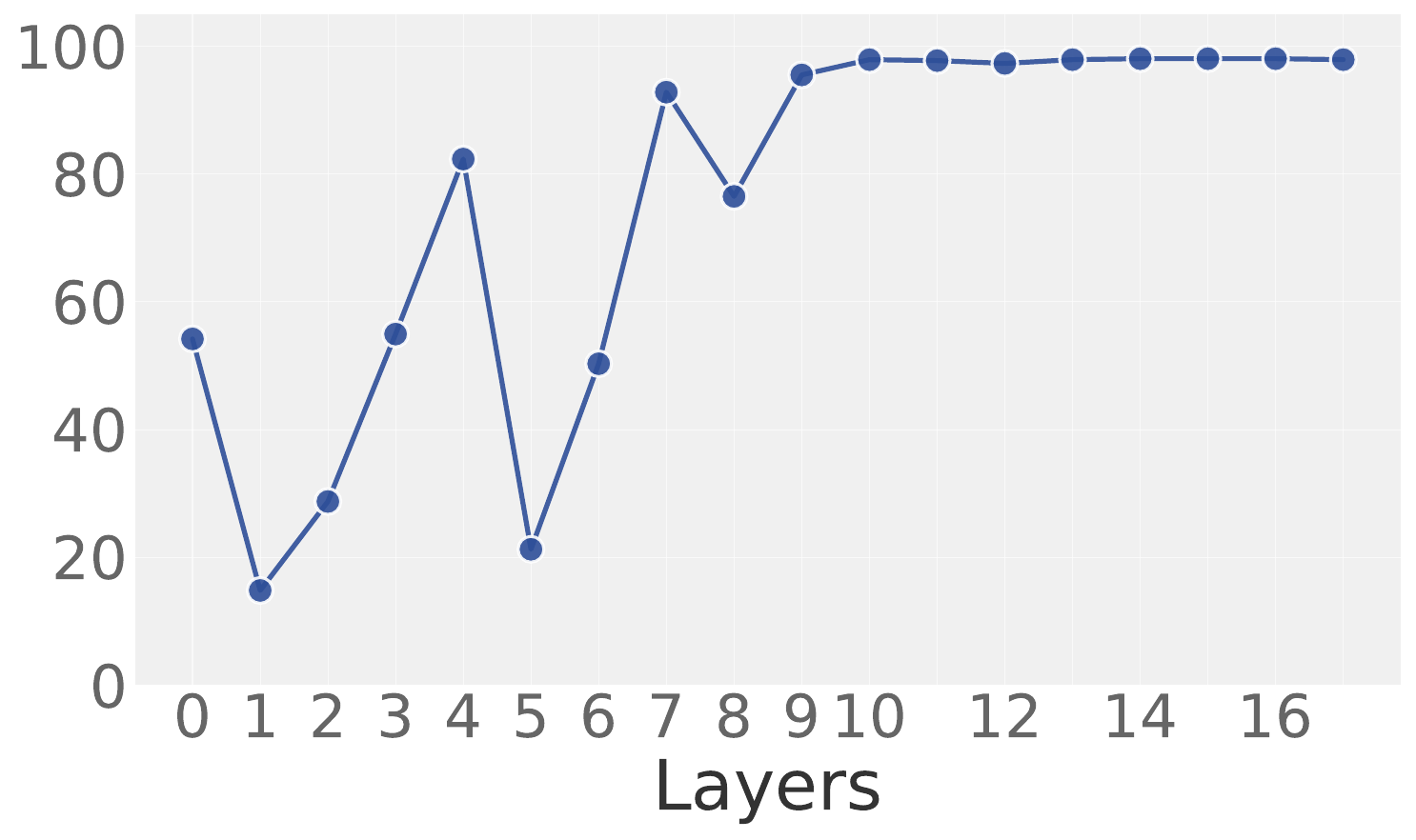}
        \caption{Gemma}
        \label{fig:zero-gemma}
    \end{subfigure}
    
    \caption{Selectively either zeroing out period and question mark tokens or non-zeroing out only these tokens for GPT-2, DeepSeek and Gemma.}
    \label{fig:zeroing-main}
\end{figure*}

\paragraph{Conditional Statements}
\label{sec:if_then}

This rule structure captures logical implication, typically in the form ``If A, then B". For example, given the base sentence \textit{``Dave is nice. If Dave is nice then he is happy."} and a override sentence \textit{``Ram is cool. If Ram is cool then he is great."}, we replace the activation of the consequent token \textit{``happy"} in the base with \textit{``great"} from the override.
We then query the model with \textit{``Is Dave great?"} to see whether it adopts the altered consequent, effectively testing whether it continues to reason over the modified rule. 

\paragraph{Universal Quantification}
\label{sec:All}

This rule type involves generalized statements about all members of a category, typically structured as \textit{``All [adjective] things are [predicate]"}. These constructions are common in both natural language and formal logic and require the model to generalize from an adjective-noun combination to a property.

To study this, we use a base sentence such as \textit{``All blue things are nice."} and a override sentence like \textit{``All green things are great."} We intervene by replacing the activation of the predicate token \textit{``nice"} in the base with the corresponding token \textit{``great"} from the override. We then query the model with {``Are all blue things great?"} to test whether it adopts the altered predicate and violates or preserves the original generalization.

\paragraph{Layer Swaps} target the activations of an entire prompt.

\section{Results}







\subsection{Punctuation Analysis}\label{sec:punctuation-results}

\paragraph{Sufficiency of Periods and Question Mark} 
In Figure \ref{fig:non-zero-gpt2-period-and-question}, we selectively keep some tokens non-zero and zero out all other tokens for GPT2. 
Here we see that for later layers (layer 7 to 11) the performance remains high even when almost all tokens are zeroed out, as long as period and question tokens are non-zero.
This means that the period and question mark tokens in these layers contain all information that is needed to answer the question, i.e. they are sufficient for answering the question.

\paragraph{Necessity of Periods and Question Mark} 
We also find in Figure \ref{fig:zero-gpt2-period-and-question} for GPT2, that when period and question mark are both zeroed out (and all other tokens are non-zero), the performance is dramatically low, which means these tokens are necessary for answering the question.

\paragraph{DeepSeek and Gemma.} For DeepSeek punctuation is sufficient in the first and last layer, and necessary in the first.
For Gemma punctuation is sufficient in layer 12, 15 and 17, but never both necessary and sufficient.

\paragraph{\emph{We further dissect these results for GPT2 below}} 

\begin{figure}[h!]
    \centering
    \begin{subfigure}[b]{0.48\linewidth}
        \includegraphics[width=\linewidth]{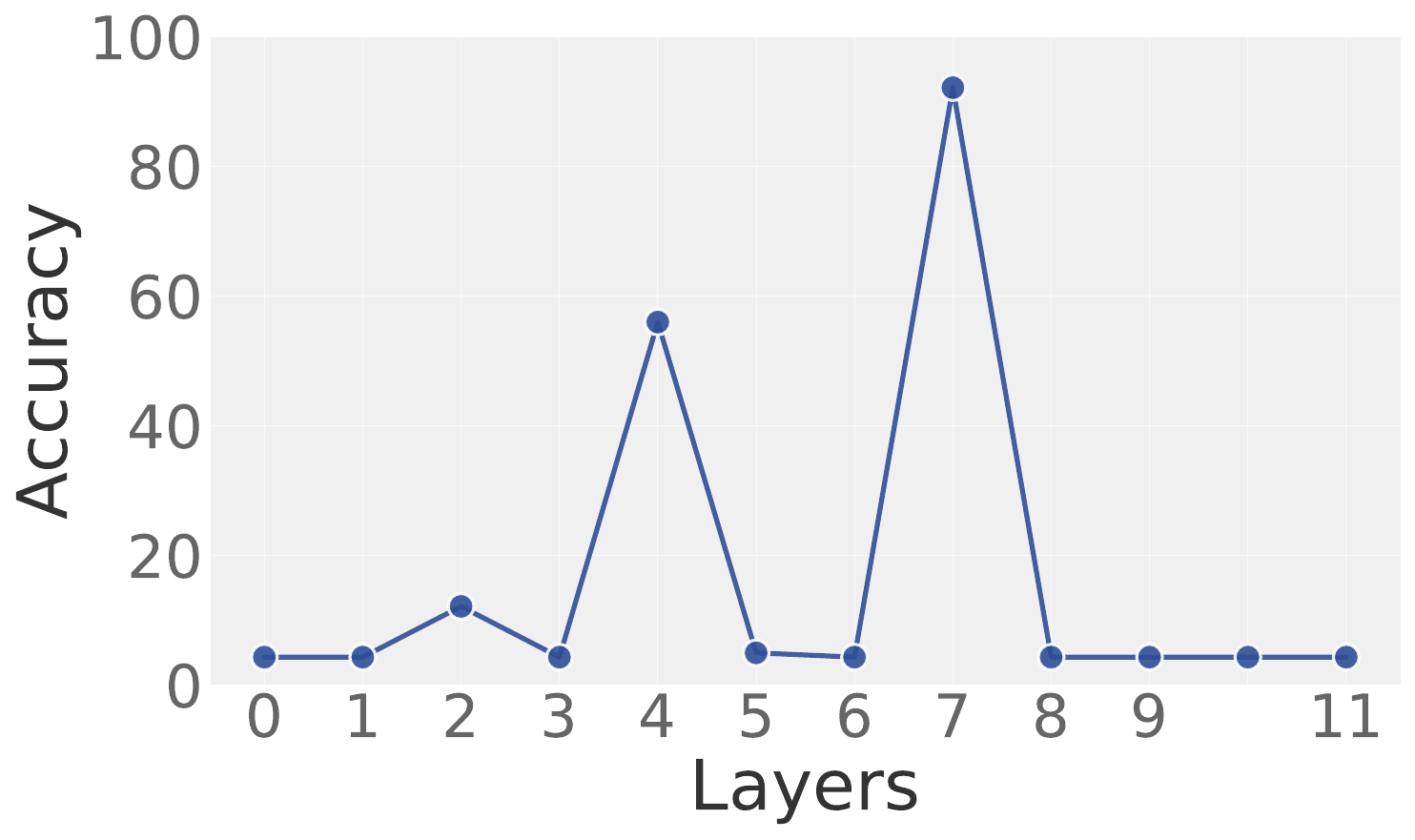}
        \caption{Periods}
        \label{fig:non-zero-gpt2-period}
    \end{subfigure}
    \hfill
    \begin{subfigure}[b]{0.48\linewidth}
        \includegraphics[width=\linewidth]{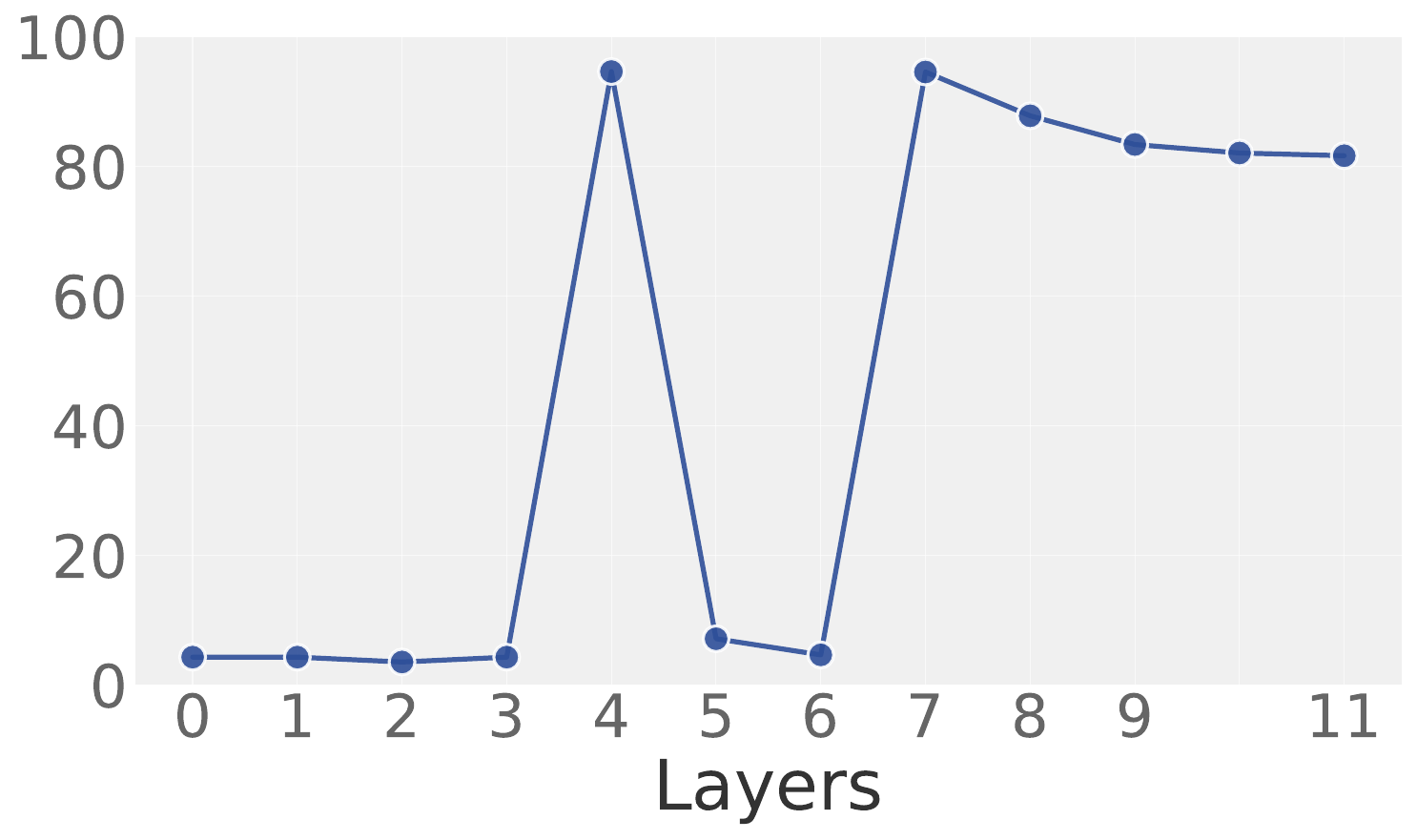}
        \caption{Question Mark}
        \label{fig:non-zero-gpt2-question}
    \end{subfigure}
    
    \vspace{0.5cm}
    \begin{subfigure}[b]{0.48\linewidth}
        \includegraphics[width=\linewidth]{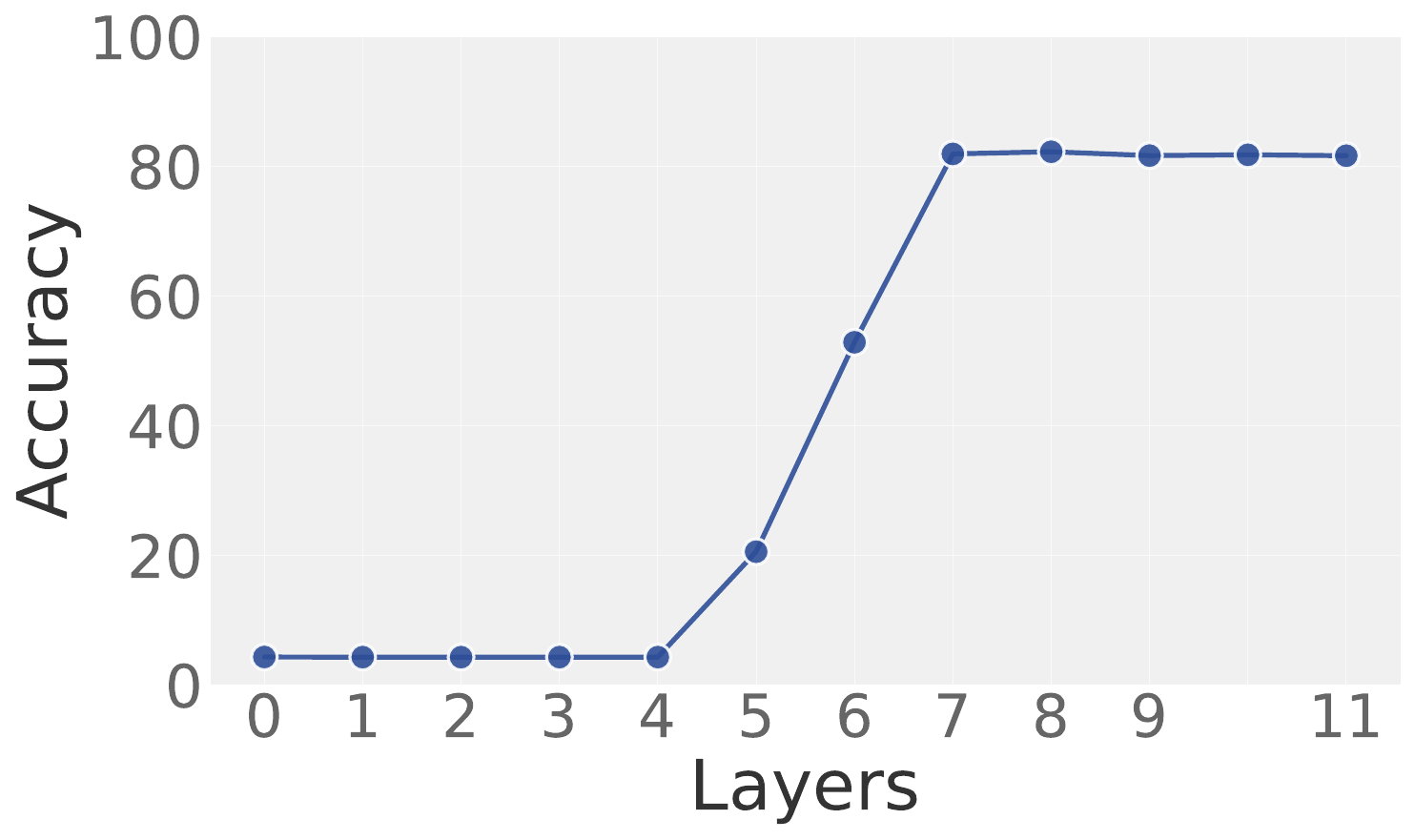}
        \caption{Periods}
        \label{fig:zero-gpt2-period}
    \end{subfigure}
    \hfill
    \begin{subfigure}[b]{0.48\linewidth}
        \includegraphics[width=\linewidth]{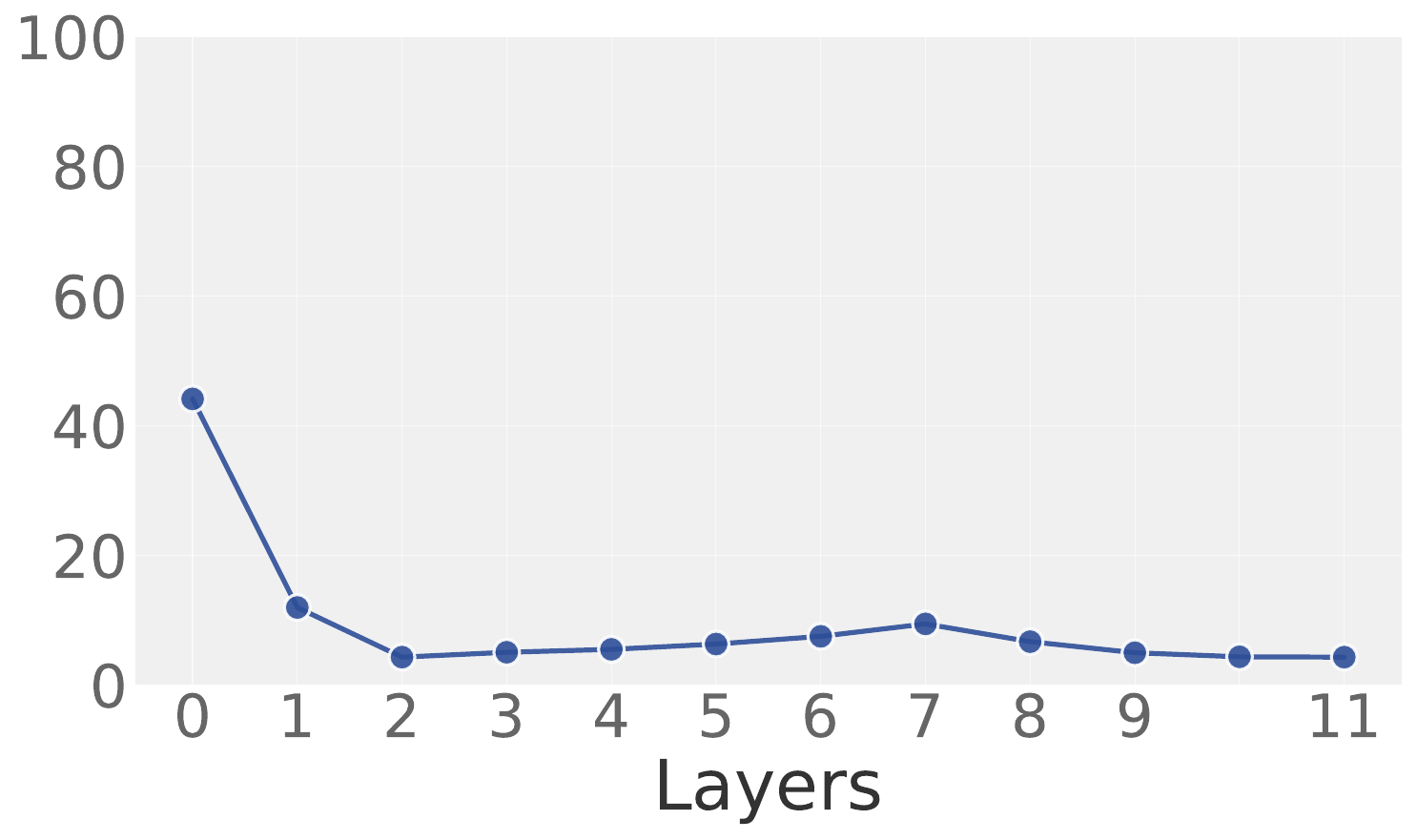}
        \caption{Question Mark}
        \label{fig:zero-gpt2-question}
    \end{subfigure}
    
    \caption{Selectively (non-)zeroing out either the period token activations or the question mark activation for GPT-2.}
    \label{fig:zeroing-dot-or-question-gpt2}
\end{figure}

\paragraph{Question Mark is Necessary and Sufficient in Some Layers}
When question mark is the only non-zero token in layer 4 or in the last five layers (Figure \ref{fig:non-zero-gpt2-question}) performance is high, but when it is the only zeroed out token in layer 4 or the last five layers 
(Figure \ref{fig:zero-gpt2-question}) the performance is low.
In contrast, period is necessary in layers 0 to 4 (Figure \ref{fig:zero-gpt2-period}), but not sufficient in those layers (Figure \ref{fig:non-zero-gpt2-period}). 

\begin{figure*}[h!]
    \centering
    \begin{subfigure}[b]{0.24\linewidth}
        \includegraphics[width=\linewidth]{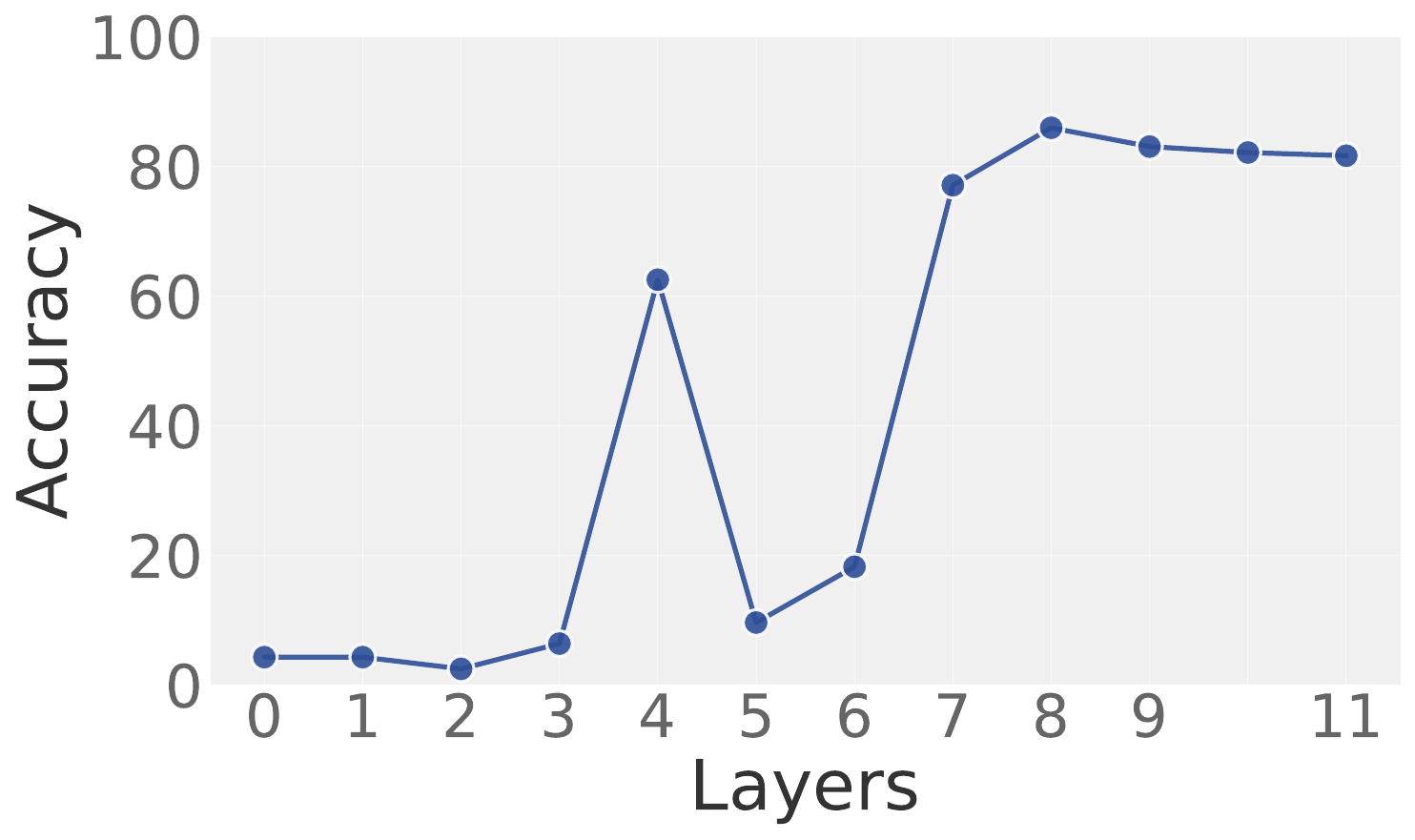}
        \caption{One in 15}
        \label{fig:1-in-15}
    \end{subfigure}
    \hfill
    \begin{subfigure}[b]{0.24\linewidth}
        \includegraphics[width=\linewidth]{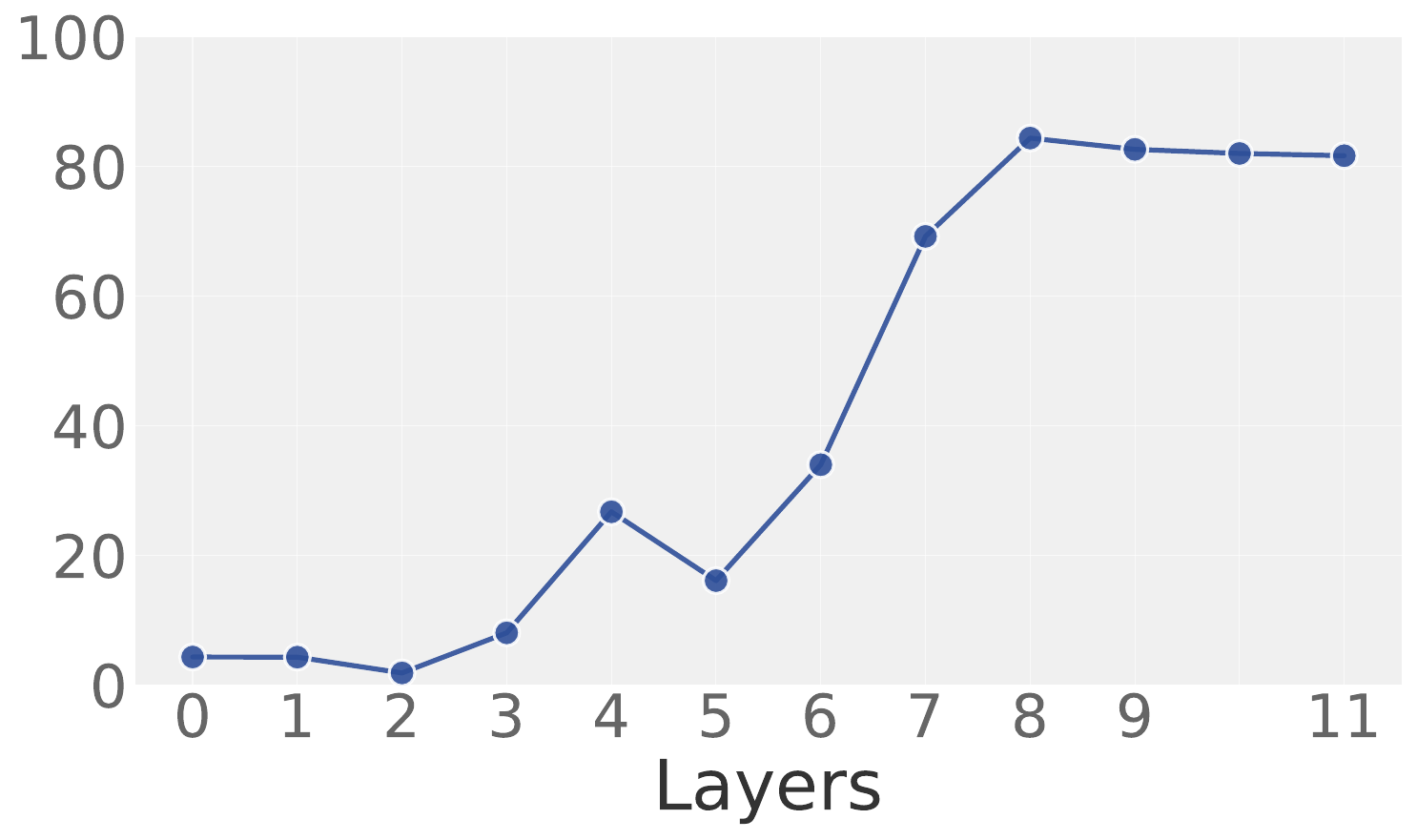}
        \caption{One in 5}
        \label{fig:1-in-5}
    \end{subfigure}
    \hfill
    \begin{subfigure}[b]{0.24\linewidth}
        \includegraphics[width=\linewidth]{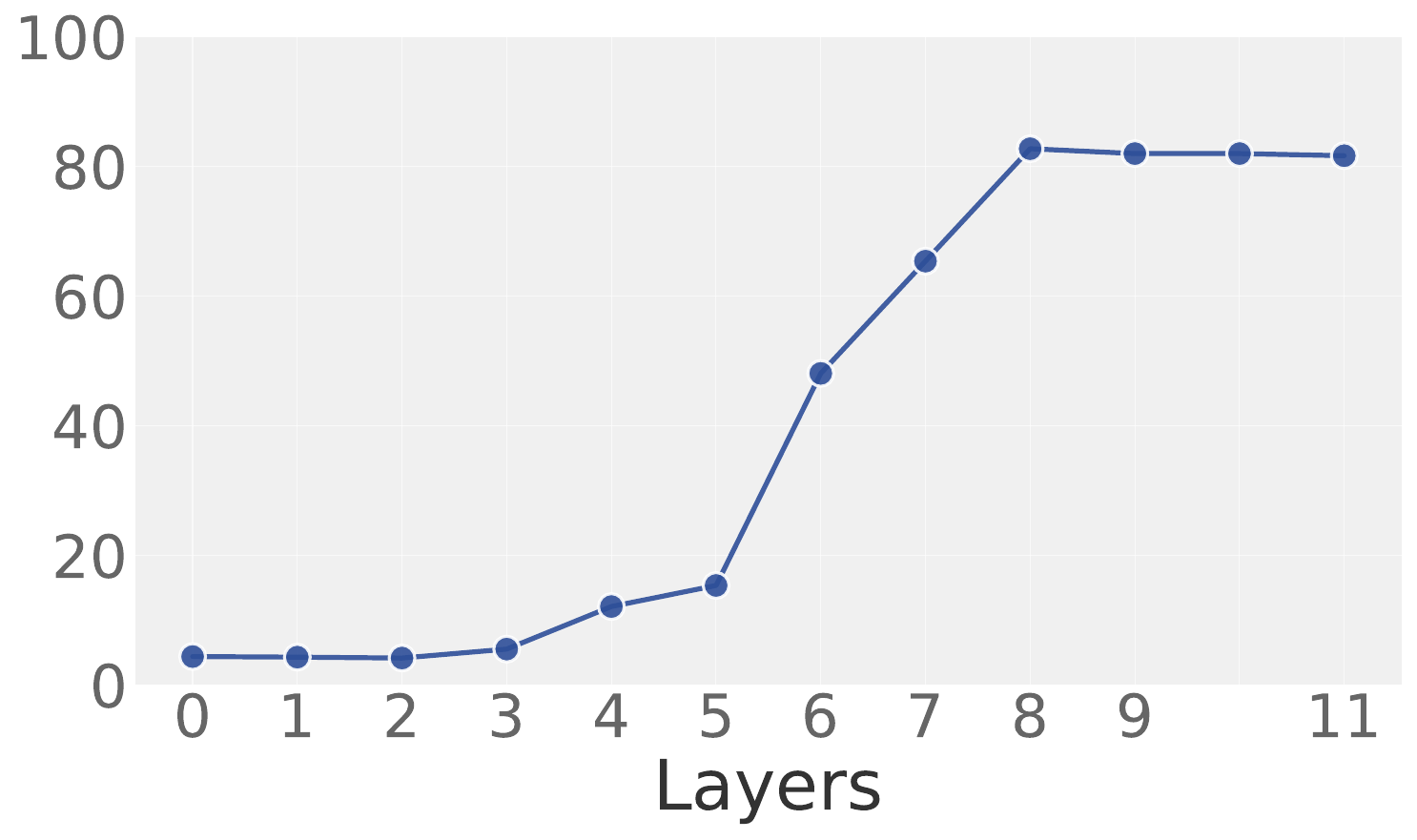}
        \caption{One in 2}
        \label{fig:1-in-2}
    \end{subfigure}
    \hfill
    \begin{subfigure}[b]{0.24\linewidth}
        \includegraphics[width=\linewidth]{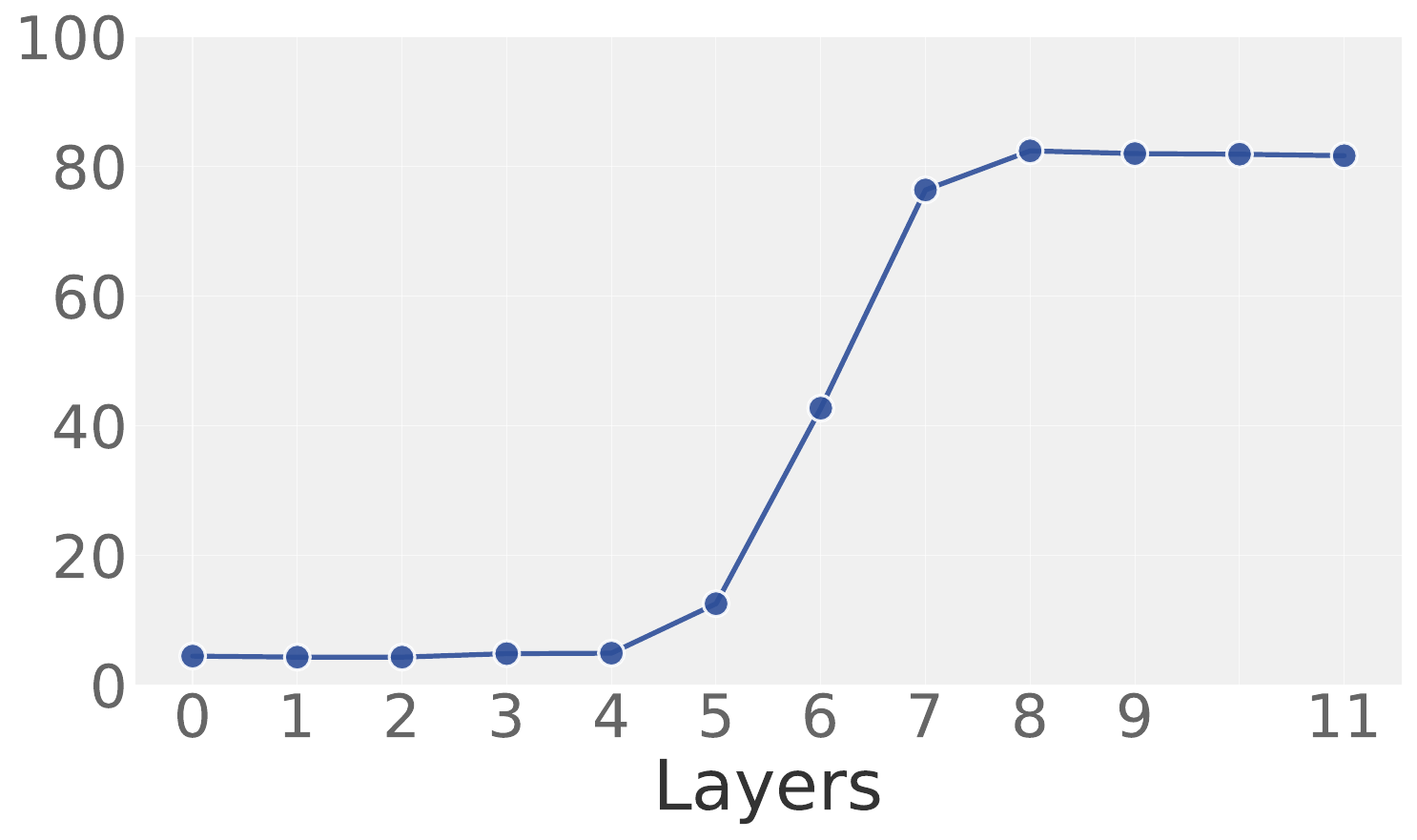}
        \caption{Four in 5}
        \label{fig:4-in-5}
    \end{subfigure}
    
    \caption{Non-zeroing question mark and different proportions of extra tokens (while keeping periods zero) for GPT-2.}
    \label{fig:nonzeroing-extra-tokens}
\end{figure*}

\paragraph{Tokens Can ``Dilute'' Question Mark}
In Figure \ref{fig:non-zero-gpt2-question} we find that if question mark is the only non-zero token in layer 4, then performance is high.
However, we also find in Figure \ref{fig:zero-gpt2-period} that if period tokens are the only zero tokens (so in particular question mark is non-zero) in layer 4 then performance is low.
We hypothesize that this is because the many other non-zero tokens in Figure \ref{fig:zero-gpt2-period} ``dilute'' the question mark token.
We investigate this hypothesis in Figure \ref{fig:nonzeroing-extra-tokens}.
In this figure question mark is always nonzero, periods are always zero, and we respectively nonzero one in 15, one in 5, one in 2, four in 5 other tokens.
We find that performance is high when question mark is one of the few nonzero tokens (see Figure \ref{fig:1-in-15}) and gradually becomes lower as we increase the proportion of nonzero tokens.

\paragraph{Transfer from Period to Question Mark}
We find that when period is non-zero in layer 7 (Figure \ref{fig:non-zero-gpt2-period}) or if question mark is non-zero in layers 7 to 11 (Figure \ref{fig:non-zero-gpt2-question}) then accuracy is high. 
We hypothesize this is because information can be ``transferred'' from period to question mark in layer 7.

\begin{figure*}[h!]
    \centering

    \textbf{First Sentence}\par\medskip
    \begin{subfigure}[b]{0.32\linewidth}
        \includegraphics[width=\linewidth]{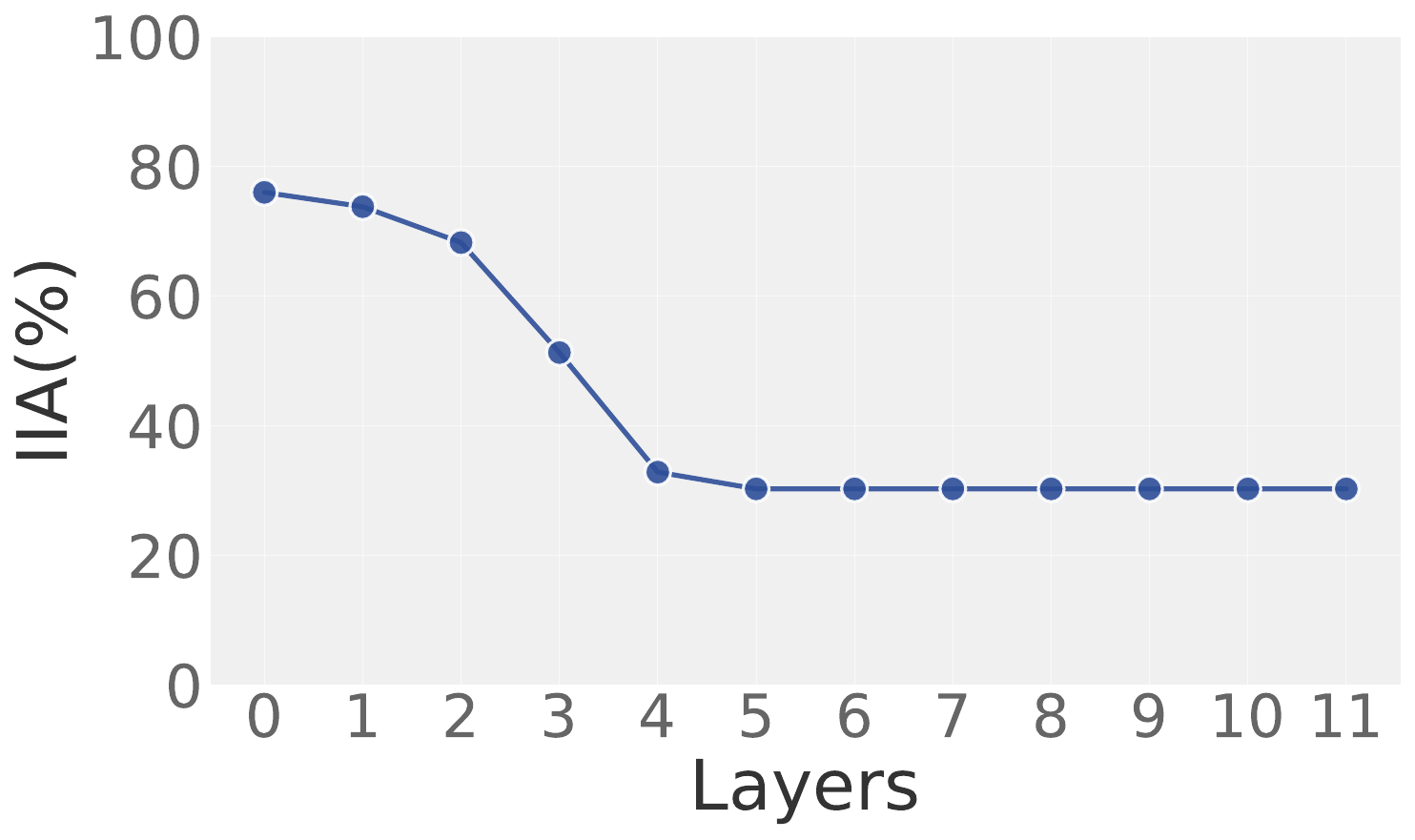}
        \caption{GPT-2}
    \end{subfigure}
    \begin{subfigure}[b]{0.32\linewidth}
        \includegraphics[width=\linewidth]{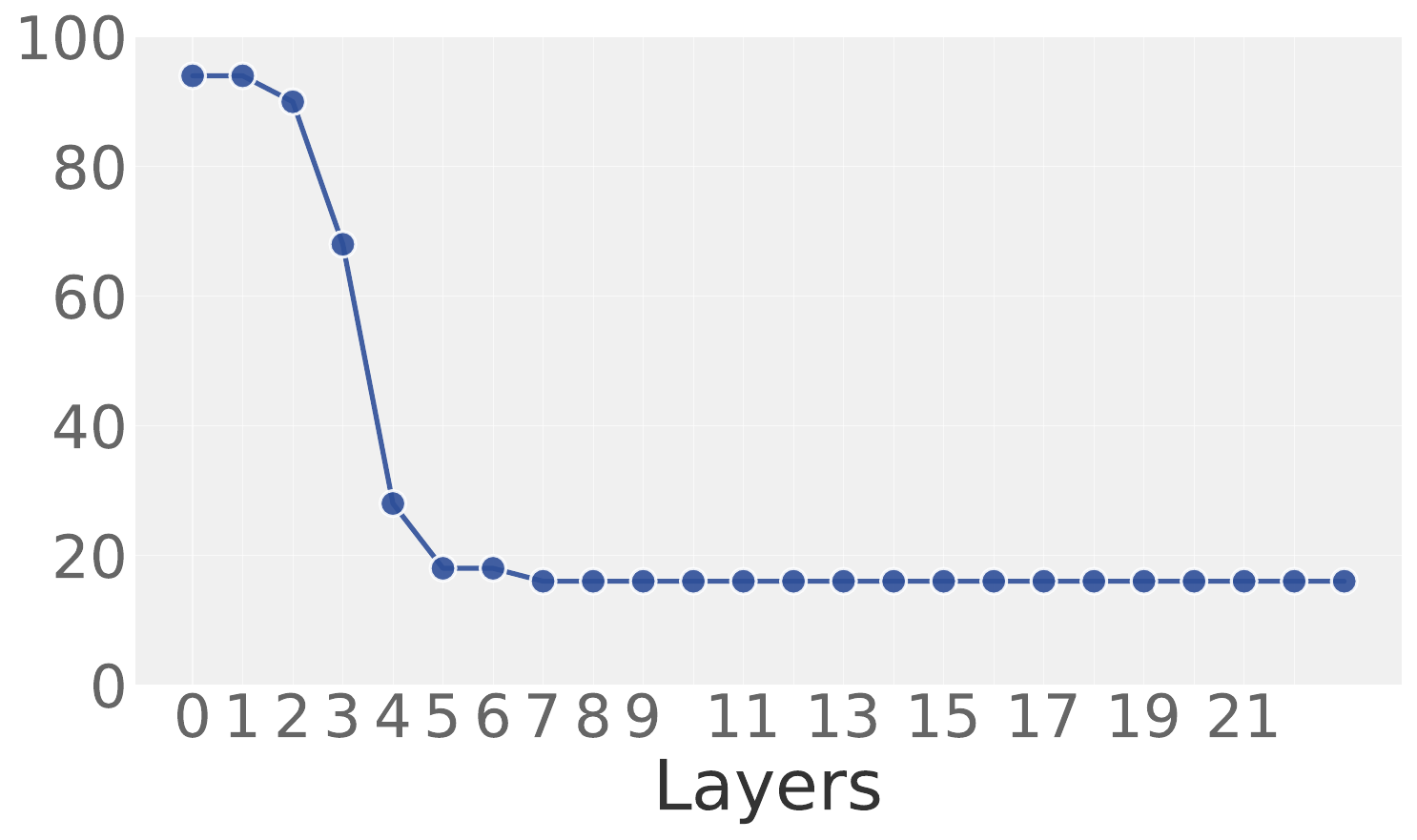}
        \caption{DeepSeek}
    \end{subfigure}
    \begin{subfigure}[b]{0.32\linewidth}
        \includegraphics[width=\linewidth]{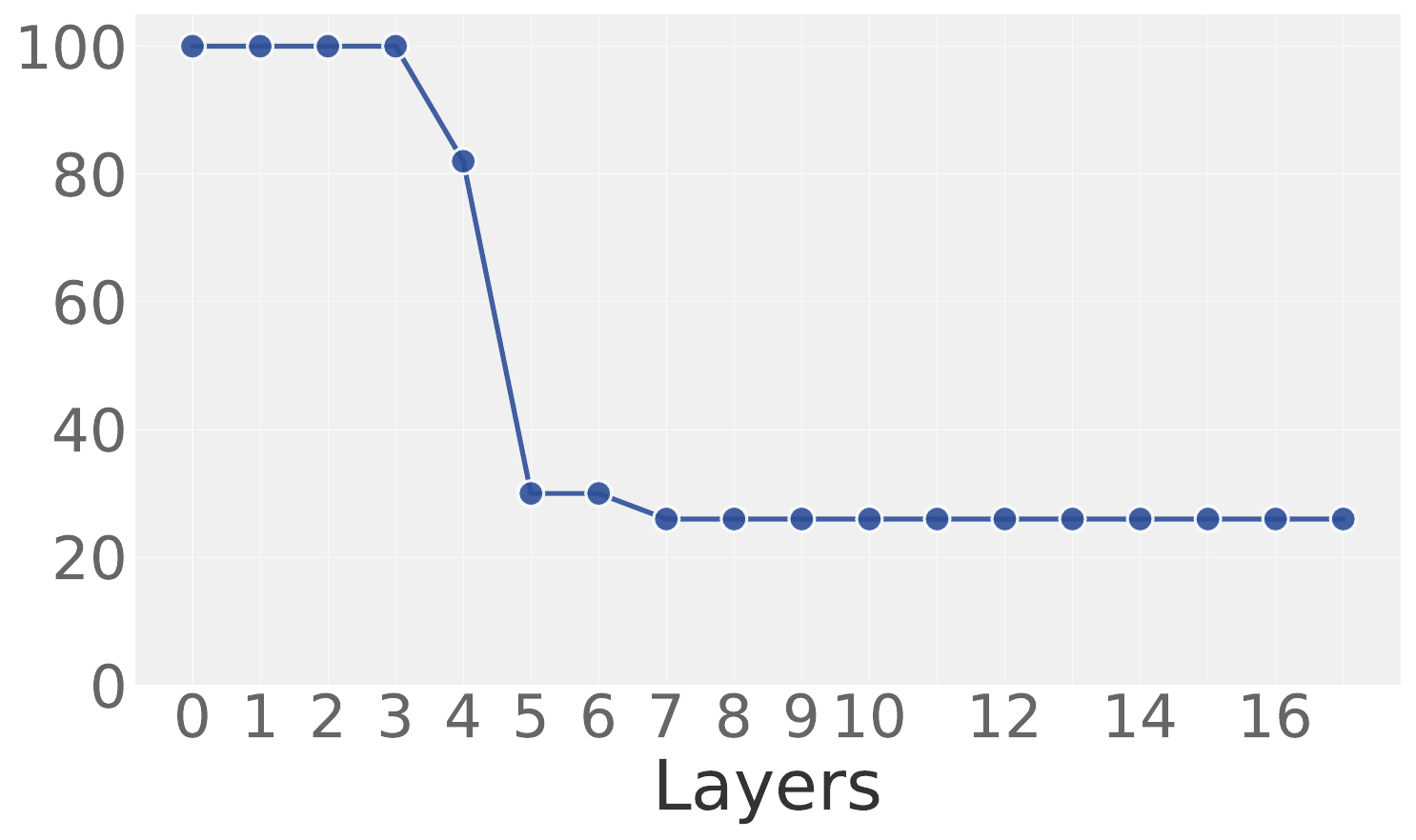}
        \caption{Gemma}
    \end{subfigure}

    \vspace{0.5cm}

    \textbf{First Period}\par\medskip
    \begin{subfigure}[b]{0.32\linewidth}
        \includegraphics[width=\linewidth]{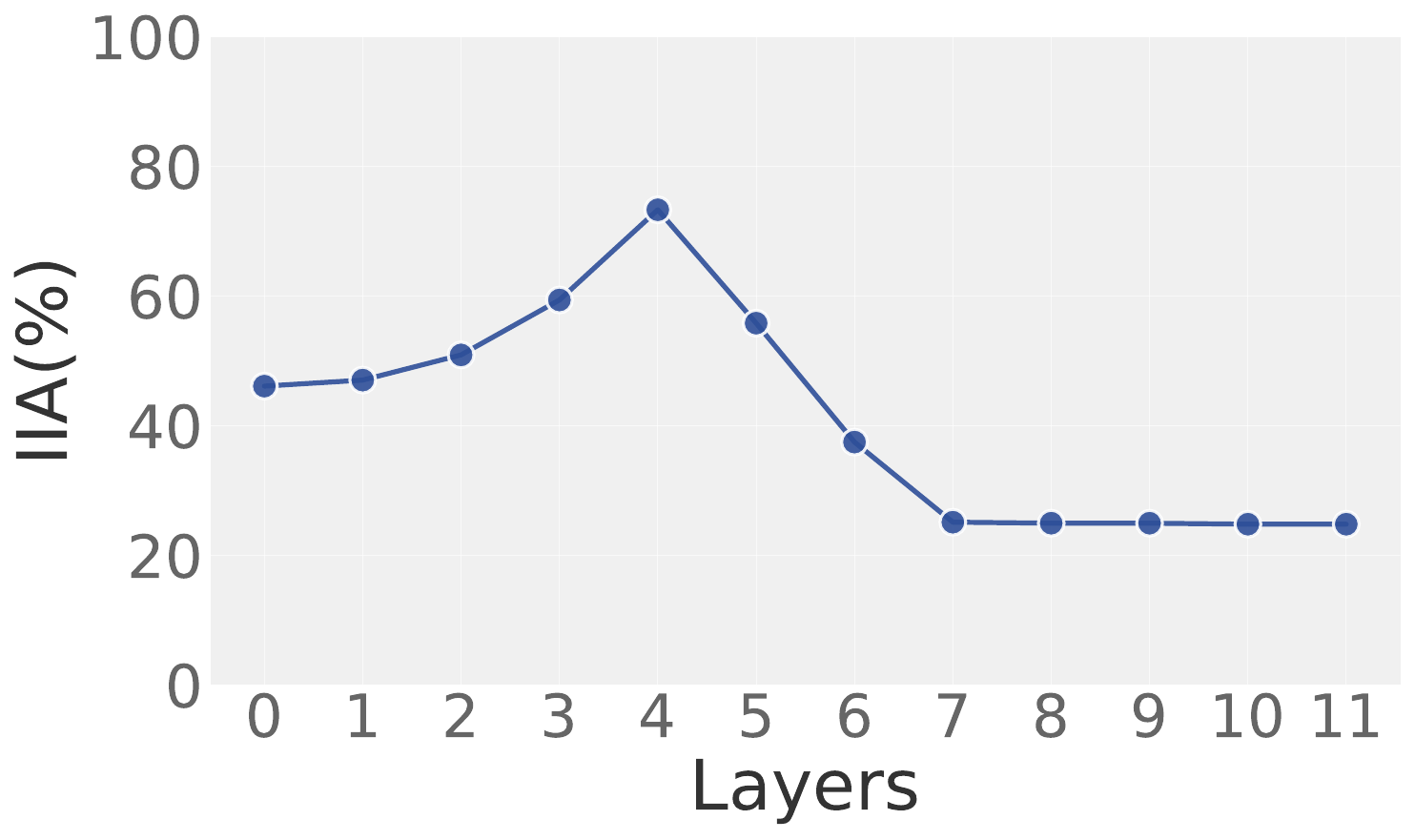}
        \caption{GPT-2}
    \end{subfigure}
    \begin{subfigure}[b]{0.32\linewidth}
        \includegraphics[width=\linewidth]{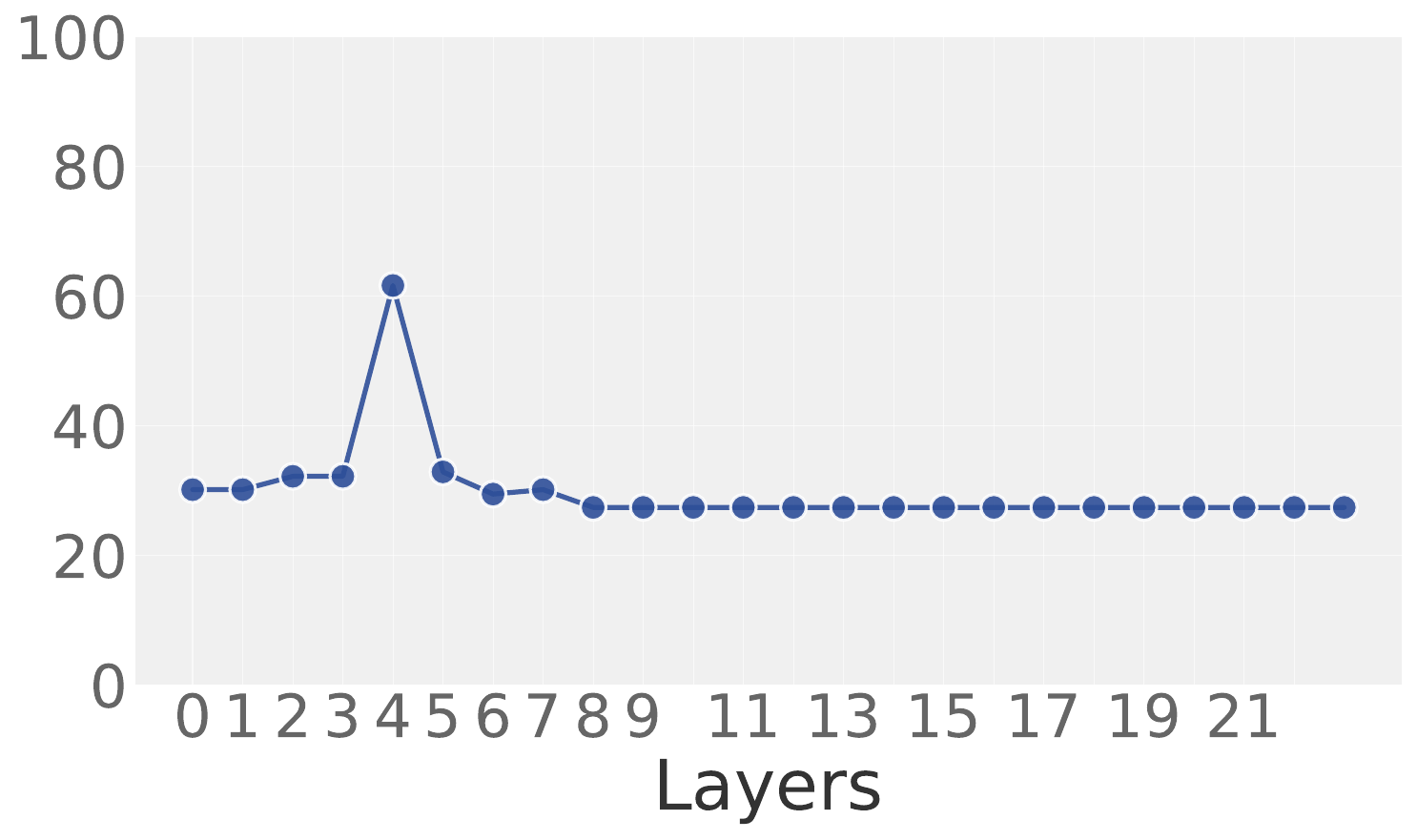}
        \caption{DeepSeek}
    \end{subfigure}
    \begin{subfigure}[b]{0.32\linewidth}
        \includegraphics[width=\linewidth]{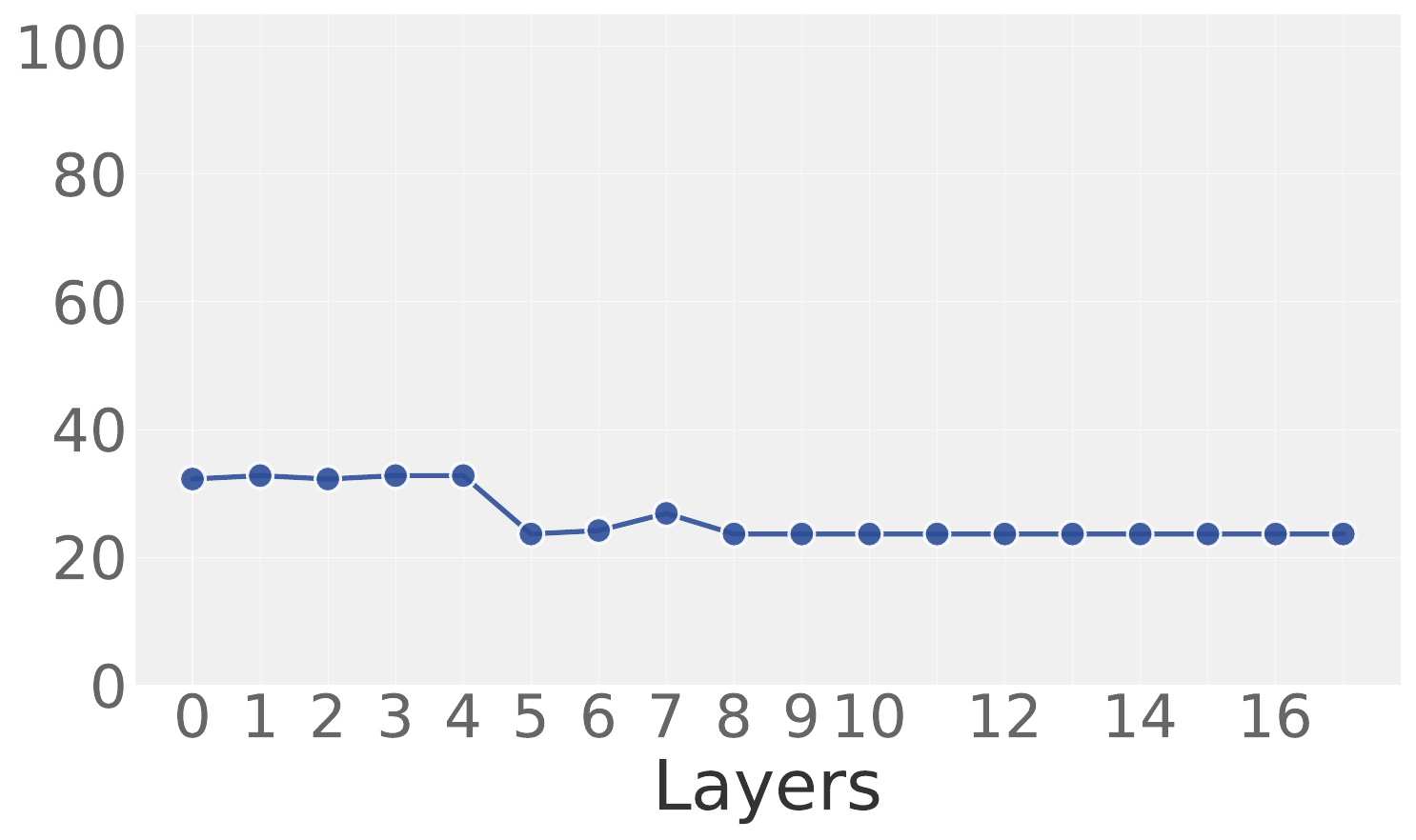}
        \caption{Gemma}
    \end{subfigure}

    \vspace{0.5cm}

    \textbf{Conditional Statements}\par\medskip
    \begin{subfigure}[b]{0.32\linewidth}
        \includegraphics[width=\linewidth]{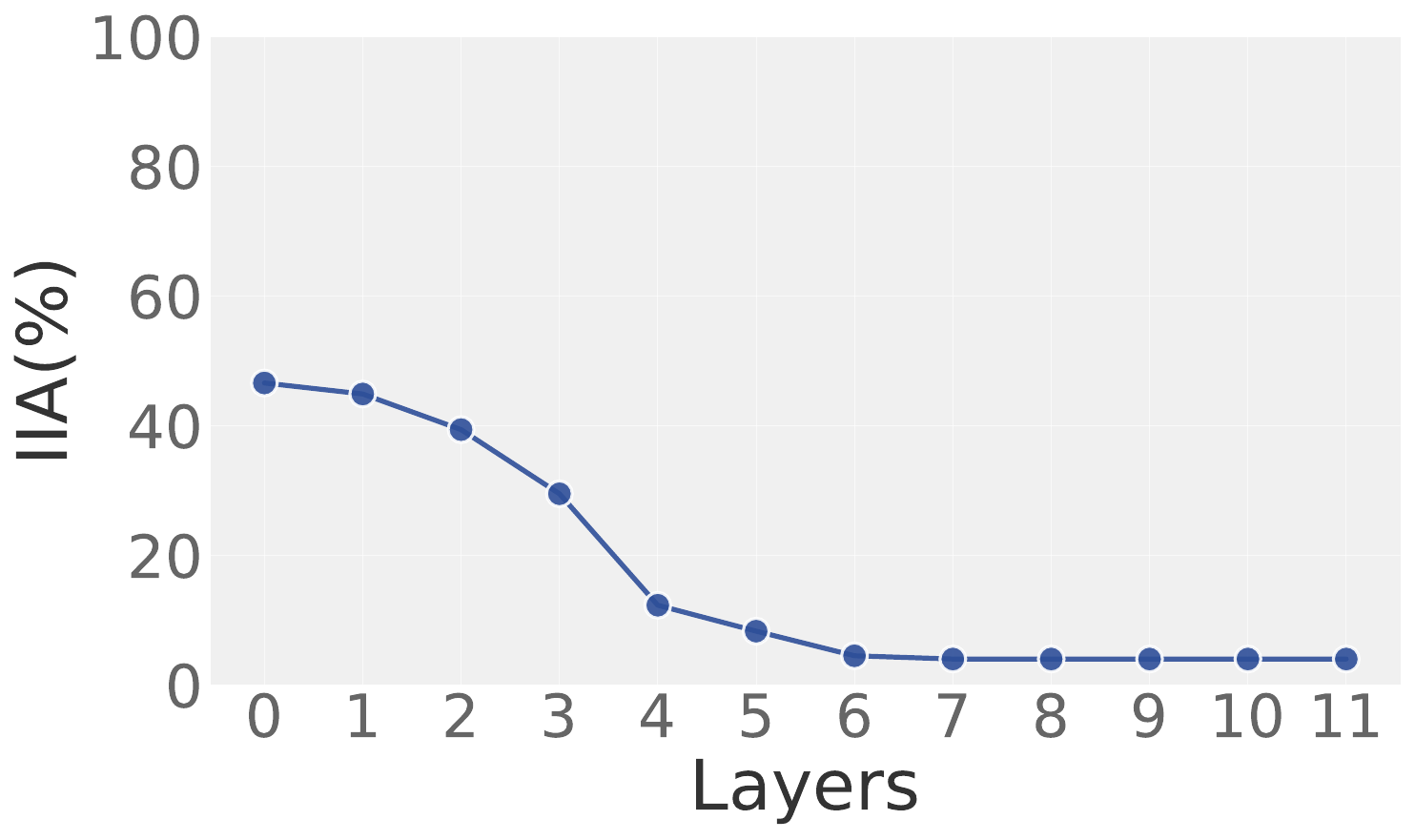}
        \caption{GPT-2}
    \end{subfigure}
    \begin{subfigure}[b]{0.32\linewidth}
        \includegraphics[width=\linewidth]{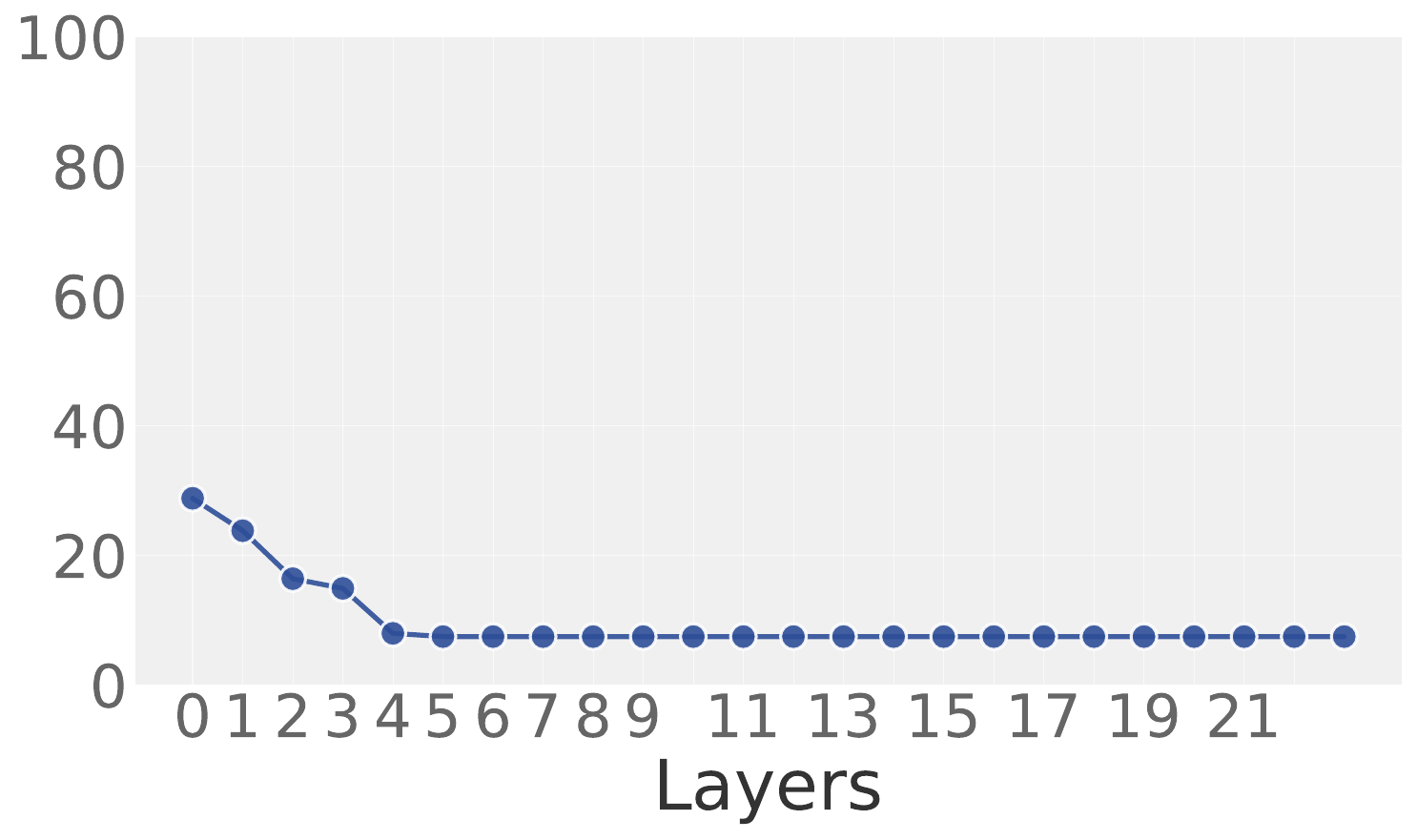}
        \caption{DeepSeek}
    \end{subfigure} 
    \begin{subfigure}[b]{0.32\linewidth}
        \includegraphics[width=\linewidth]{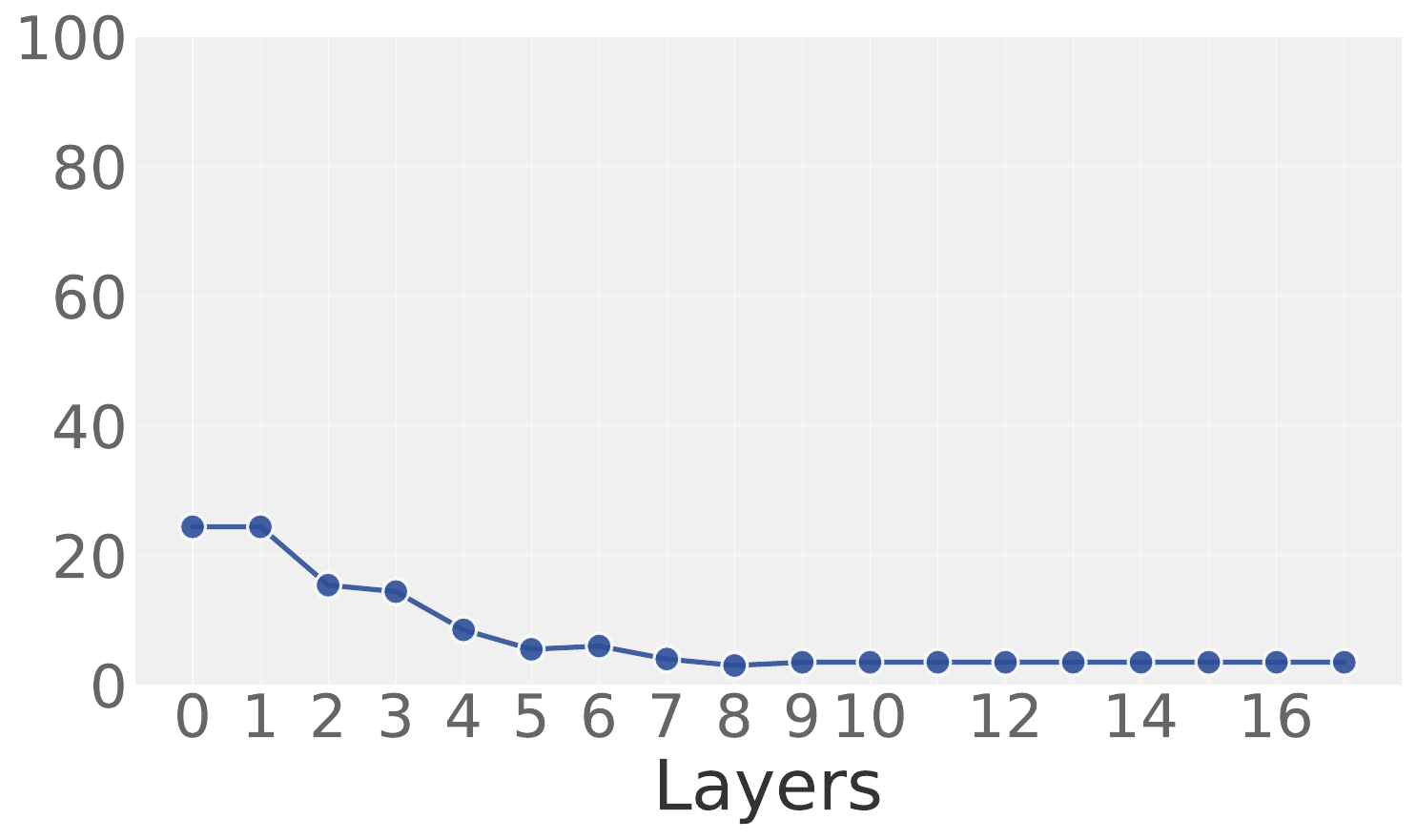}
        \caption{Gemma}
    \end{subfigure}
    
    \vspace{0.5cm}

    \textbf{Universal Quantification}\par\medskip
    \begin{subfigure}[b]{0.32\linewidth}
        \includegraphics[width=\linewidth]{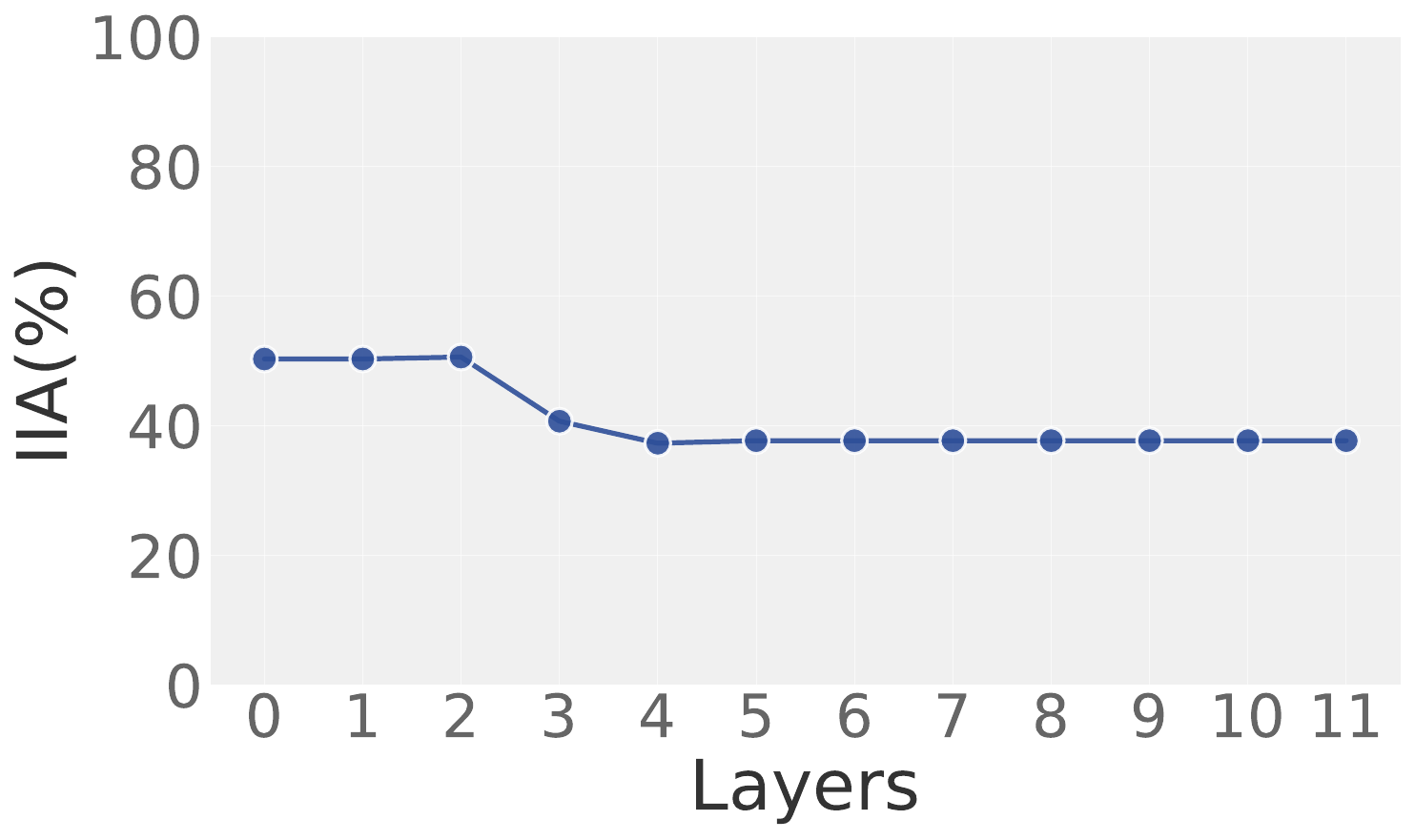}
        \caption{GPT-2}
    \end{subfigure} 
    \begin{subfigure}[b]{0.32\linewidth}
        \includegraphics[width=\linewidth]{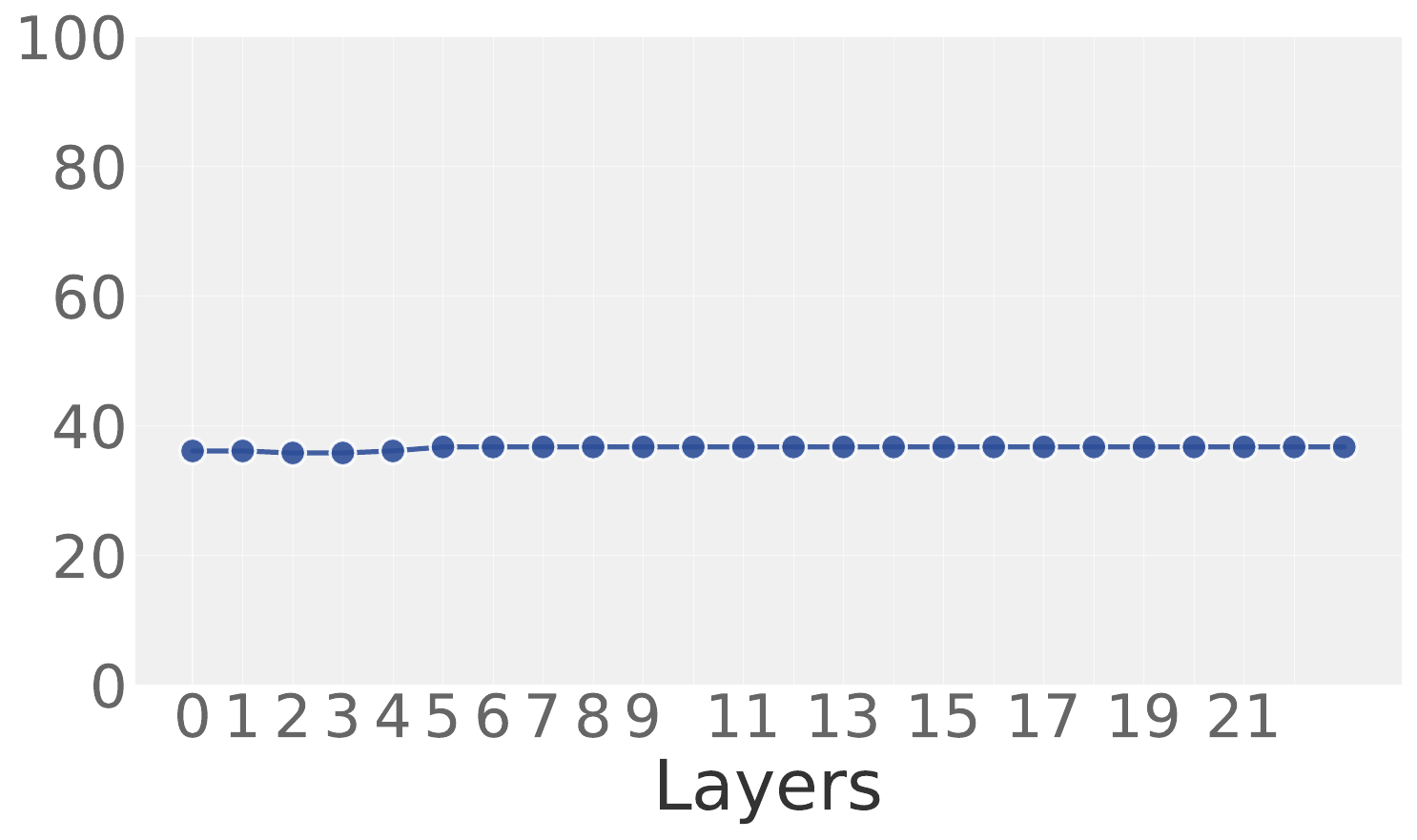}
        \caption{DeepSeek}
    \end{subfigure}
    \begin{subfigure}[b]{0.32\linewidth}
        \includegraphics[width=\linewidth]{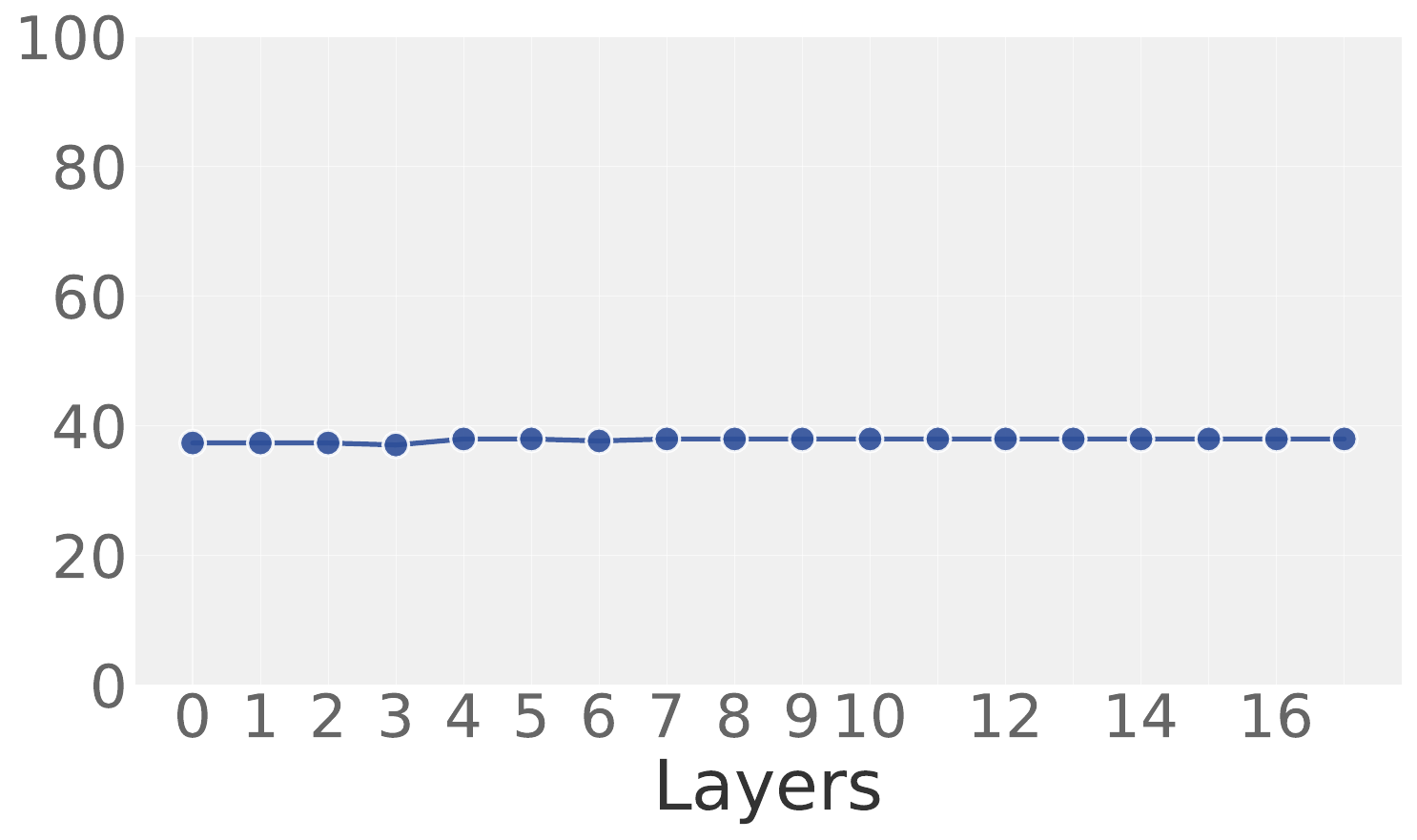}
        \caption{Gemma}
    \end{subfigure}

    \caption{Layer-wise sensitivity of GPT-2, DeepSeek, and Gemma to interchange interventions applied to different intervention targets 
    (first sentence, first period, conditional statements, universal quantification). 
    Each row represents an intervention target; each column corresponds to a model.
    When applying interchange intervention we take a sentence and replace a sentence component $z$ with an alternative $z'$ and check whether the model's response aligns with the response we would expect if $z'$ was in the prompt, if so we say the IIA for this datapoint is 1, if not it is 0.
    See the Evaluation Metrics section for more details on how IIA is calculated and the Intervention Targets section for more details on the intervention targets. 
    }
    \label{fig:interchange-intervention-results}
\end{figure*}

\paragraph{First Full Sentence Interchange Intervention Accuracy (IIA)} 
For all three models we find in Figure \ref{fig:interchange-intervention-results} that when we perform interchange intervention on the first full sentence of the prompt, the IIA starts out high, and then dramatically decreases.

\paragraph{First Period IIA}
The results shown in Figure \ref{fig:interchange-intervention-results}, also indicate that for GPT-2 the IIA peaks in layer 4. 
DeepSeek shows a similar but more abrupt pattern, starting out at an IIA of just above 20\%, remaining constant, suddenly peaking at layer 4 to around 60\%, and then dropping to 20\% again.
Gemma, in contrast, shows negligible sensitivity to interchange interventions for first full period.
In line with previous findings  \cite{barbero2025llmsattendtoken}, we hypothesize that this difference is related to the context window size of the models: GPT-2 has the smallest window, followed by DeepSeek and then Gemma, or due to Gemma being a distilled model. 


\paragraph{First Sentence IIA Drop Coincides with First Period Peak for GPT2 and DeepSeek}
For GPT-2 the layer for which the IIA of the first period peaks coincides with the layer for which the IIA of the first sentence finishes its drop.
The peak for first period at layer 4 for DeepSeek again coincides with a large drop for first sentence at layer 4.
We hypothesize that in the first four layers information from the first sentence is ``transferred'' to the first period.
This is corroborated by Figure \ref{fig:zero-gpt2-period}, where we find that periods contain important information in the first four layers.
However, we would have expected the IIA to either be constantly high, or the zeroing out accuracy to gradually decrease across the first four layers.
In Figure \ref{fig:additional-IIA-results} in the appendix we find a similar pattern for second sentence and second period, but with a lower final IIA.

\subsection{Reasoning Rules}\label{sec:reasoning-results}


Based on previous work \cite{sun2025transformerlayerspainters, yang2025internalchainofthoughtempiricalevidence} and before starting interchange intervention experiments on reasoning rules we had expected to see the model collecting information in early layers and applying reasoning rules in later layers.
However, for GPT-2 our results instead show that similar to first sentence, second sentence, adjective and subject interventions (Figure \ref{fig:interchange-intervention-results} and Figure \ref{fig:additional-IIA-results} in the appendix) intervening on reasoning rules leads to high IIA in the early layers and low IIA in later layers.

\paragraph{Conditional Statements} 
In Figure \ref{fig:interchange-intervention-results} we plot the IIA for conditional rules. 
This means that we take a sentence with ``If $x$, then $z$" and replace the activations of the consequent $z$ with the activations of $z'$.
We find that GPT-2 has a high IIA in the first layer of around 50\%, but this has drastically dropped by layer 6, where the IIA is around 5\%, and the IIA remains low for all following layers. 
We interpret this as the model actively processing information about the consequent ($z$ or $z'$) in the first 5 layers, but after that (from layer 6 onward), the activations in previous layers already determine which consequent the model works with.
For DeepSeek and Gemma we find a similar pattern, but here the IIA respectively starts around 30\% and 25\%, and respectively drops around layer 4 and layer 7.
We find that proportionally DeepSeek processes the consequent very quickly, namely in 4 layers (out of 22), whereas GPT-2 has only completed processing the consequent after the middle layer.


\begin{figure*}[h]
    \centering

    \textbf{Conditional Statements}\par\medskip
    \begin{subfigure}[b]{0.32\linewidth}
        \includegraphics[width=0.7\linewidth]{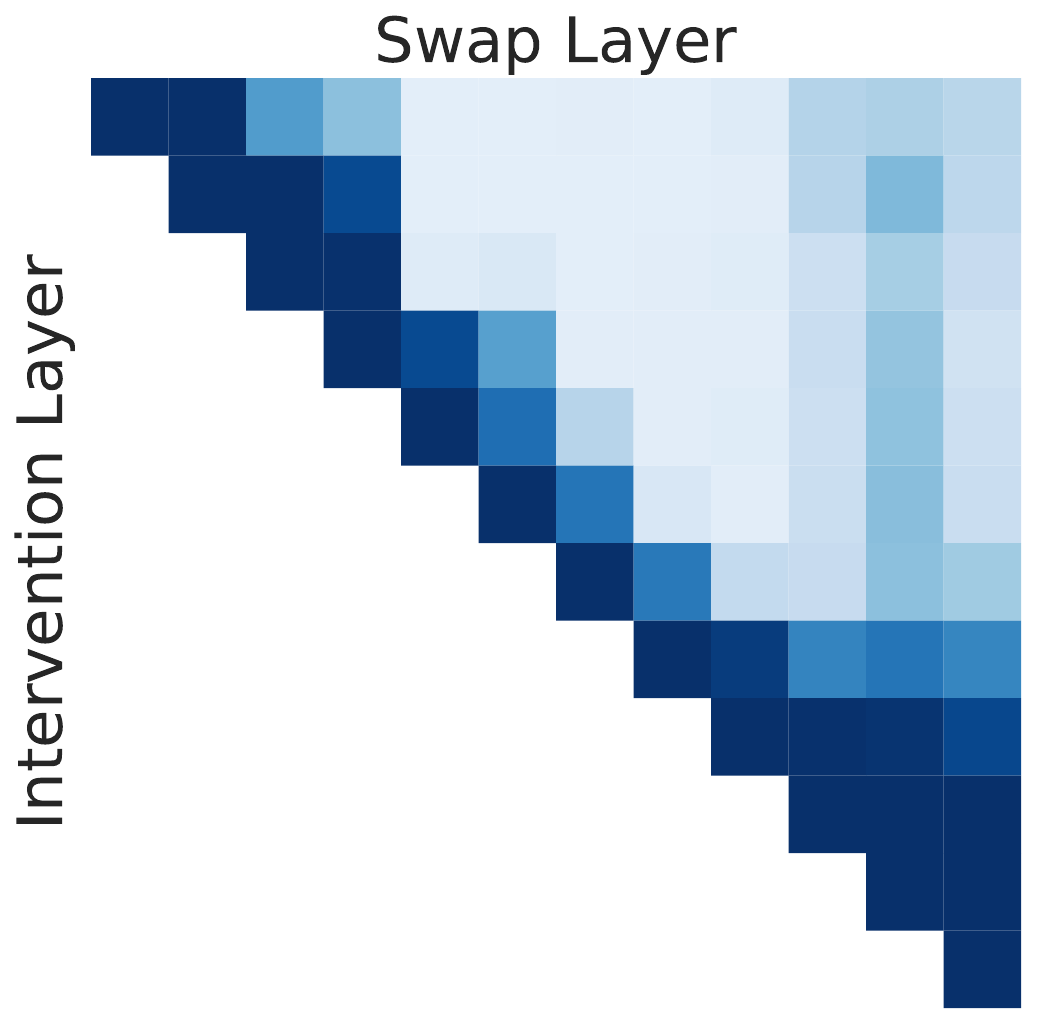}
        \caption{GPT-2}
    \end{subfigure}
    \begin{subfigure}[b]{0.297\linewidth}
        \includegraphics[width=0.7\linewidth]{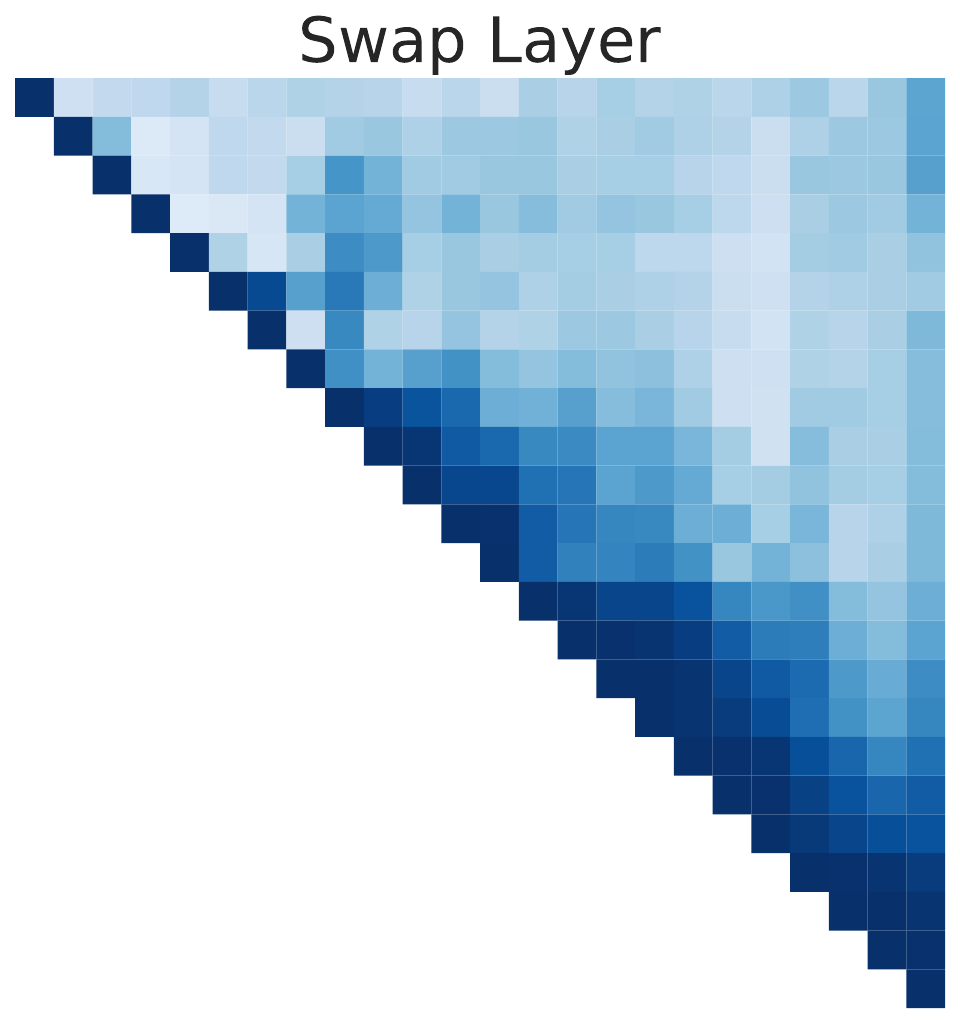}
        \caption{DeepSeek}
    \end{subfigure}
    \begin{subfigure}[b]{0.342\linewidth}
        \includegraphics[width=0.7\linewidth]{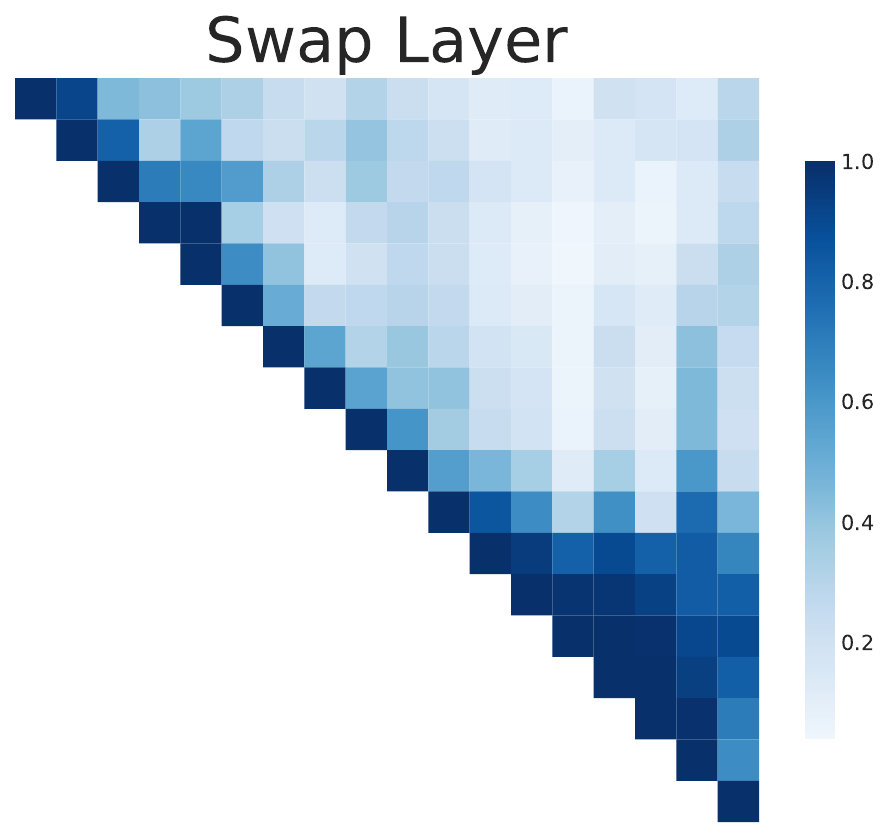}
        \caption{Gemma}
    \end{subfigure}

    \vspace{0.5cm}

    \textbf{Universal Quantification}\par\medskip
    \begin{subfigure}[b]{0.32\linewidth}
        \includegraphics[width=0.7\linewidth]{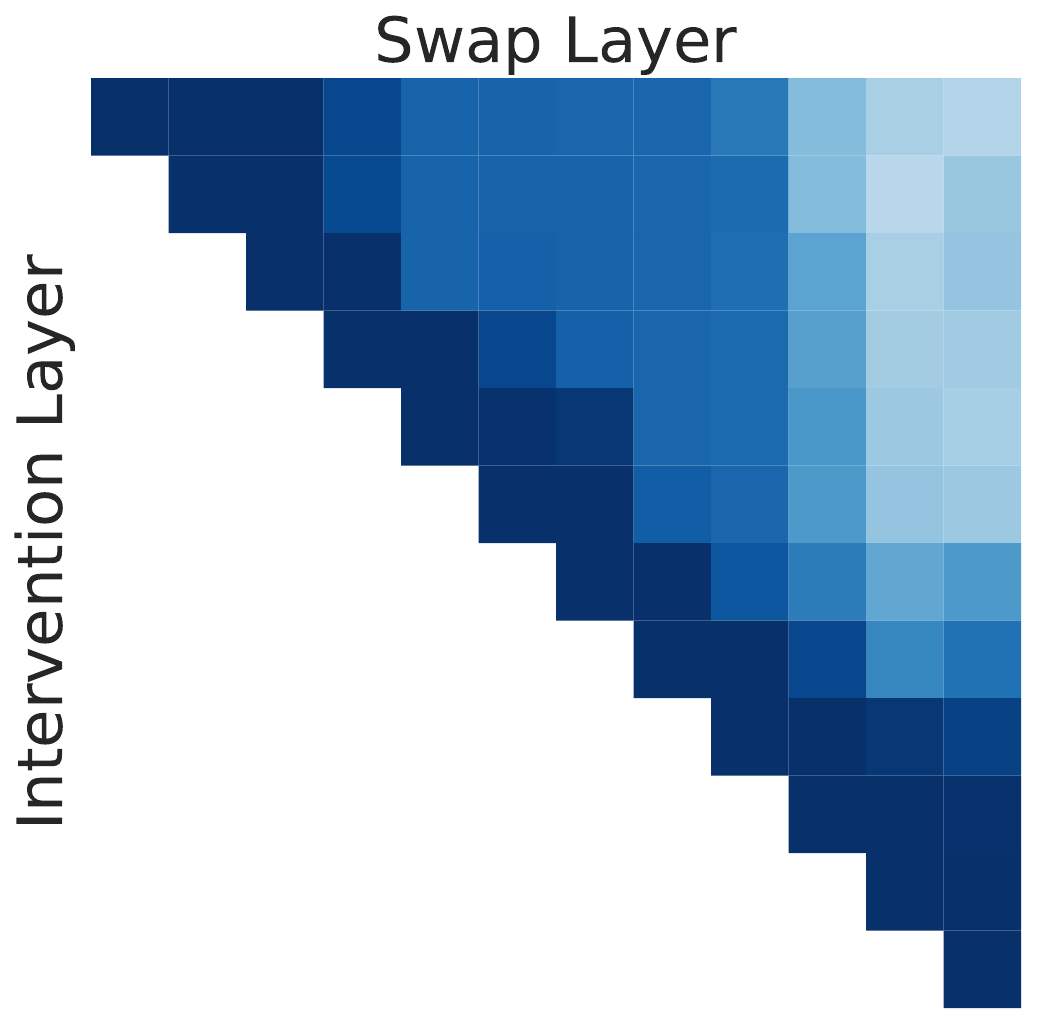}
        \caption{GPT-2}
    \end{subfigure}
    \begin{subfigure}[b]{0.297\linewidth}
        \includegraphics[width=0.7\linewidth]{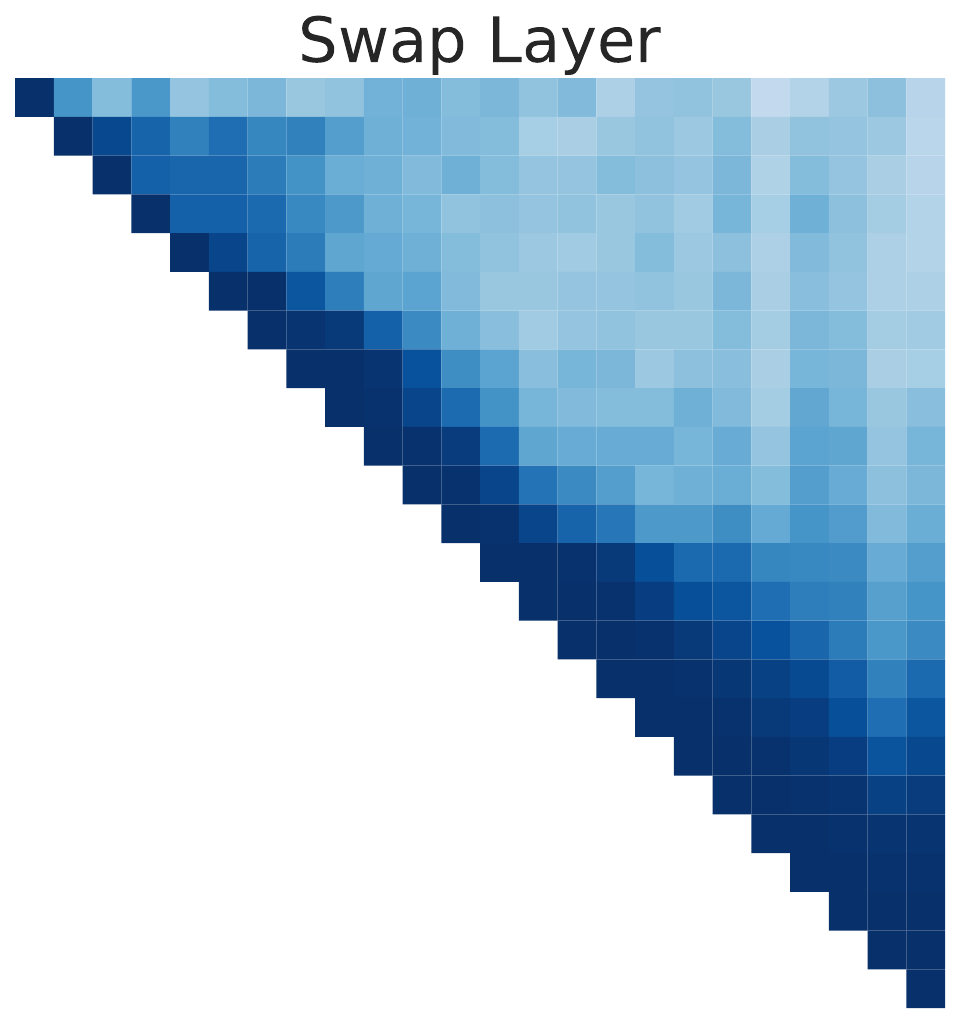}
        \caption{DeepSeek}
    \end{subfigure}
    \begin{subfigure}[b]{0.342\linewidth}
        \includegraphics[width=0.7\linewidth]{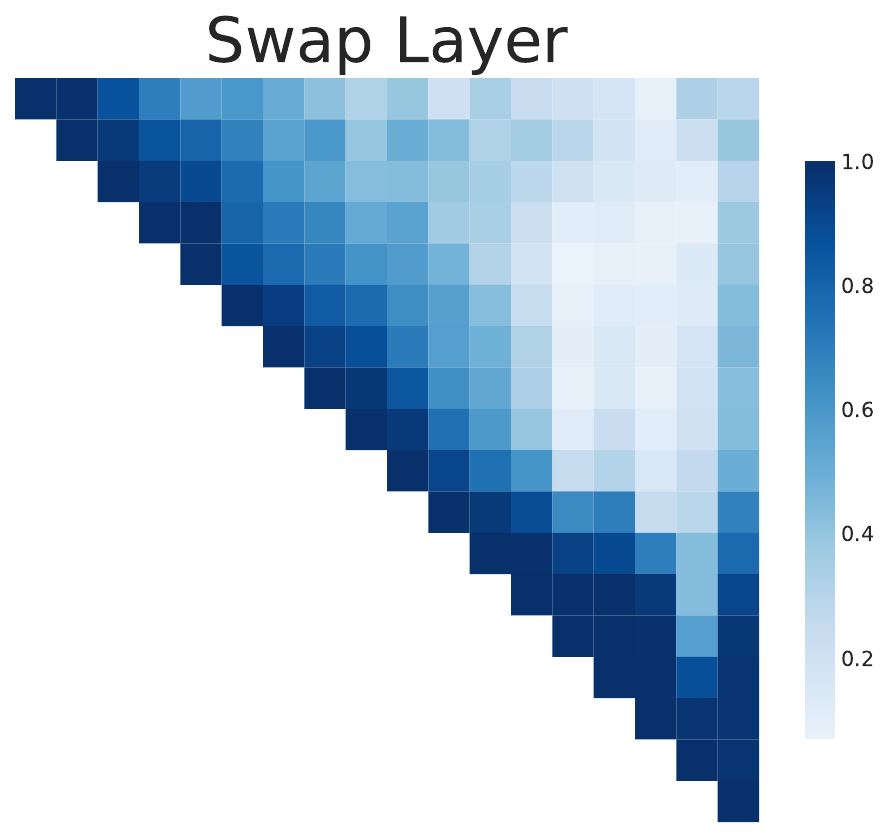}
        \caption{Gemma}
    \end{subfigure}

    \caption{Comparison of layer swap heatmaps across different language models. Each heatmap shows the correlation between layers when performing layer swapping experiments. Higher correlation values indicate that swapping those layers has minimal impact on model performance, while lower values suggest significant performance degradation.}
    \label{fig:all_heatmaps}
\end{figure*}

\paragraph{Universal Quantification}
We plot the IIA for universal quantification in Figure \ref{fig:interchange-intervention-results}.
In a sentence like ``All $x$ are $z$", we replace the predicate $z$ with $z'$.
The surprising finding here is that for all models, the lowest IIA is still very high, around 40\%.
This means that for all layers replacing the predicate $z$ with $z'$ leads to the model answering as if the prompt said $z'$ in 40\% of the datapoints. 
In other words, the model does not ``stop processing'' the predicate after some layer (which is unlike how the models handled conditional rules). 

\paragraph{IIA Drops to Very Low Point for Conditional Statements}
Comparing adjective, subject, universal quantification and conditional statements, 
we find that conditional statements is the only intervention where the IIA drops to a very low point.
Out of these universal quantification is the only target where the the IIA is constant for two models (DeepSeek and Gemma).
We interpret this as the model revisiting the universal quantification in equal measure in every layer.



\paragraph{Layer Swaps}
In Figure \ref{fig:all_heatmaps} each entry represents the impact of a layer swap, where the x-axis represents the the layer we intervene on, and the y-axis represents layer we swap it with.
The swap layer plot, see Figure \ref{fig:all_heatmaps}, indicates that for conditional statements layers have specific functions, and can not replace each other, whereas for universal quantification layers are more replaceable. 
We speculate this may indicate that conditional statements are more difficult for the model to solve as compared to universal quantification.






\section{Discussion}

We analyzed the importance of `.' and `?' through the lens of necessity and sufficiency of these tokens for model performance. 
We shed light on the reasoning process of models and evaluate whether models process different reasoning rules differently.

\subsection{Limitations and Future Work}

We only do necessity and sufficiency investigations for punctuation. 
Future work could expand this investigation to other tokens.
Additionally, we only investigate periods and question mark in isolation for GPT-2, this investigation could be expanded to a wider variety of models.

In intervention analysis we only did token level interventions per layer, it would be better to have more finegrained intervention and subspace analysis. Future work could do a deeper subspace analysis.

In our reasoning investigation we did part-of-sentence interventions, where we targeted the consequent or predicate in a prompt.
Although we did find differences in IIA between different part-of-sentence targets, it is hard to know to what extent our results apply to syntactic processing of tokens in general, or to reasoning specifically.
Future work could investigate swapping entire rules (for example ``if then'' statements, with ``if and only if then'' statements).

\subsection{Acknowledgments}

We would like to thank AI Safety Camp for hosting and supporting this project, and for providing essential compute resources. Sonakshi Chauhan is grateful to Atticus Geiger for his valuable research guidance, feedback, and support during the initial development of this project.


\bibliography{references}


\section{Appendix}
\label{sec:appendix}

\vspace{1cm}

\begin{center}
\footnotesize
\setlength{\tabcolsep}{3pt}
\renewcommand{\arraystretch}{1.1}
\begin{tabular}{@{}p{0.2\textwidth}p{0.24\textwidth}p{0.24\textwidth}p{0.24\textwidth}@{}}
\toprule
\textbf{Feature} & \textbf{GPT2-small} & \textbf{DeepSeek} & \textbf{Gemma} \\
\midrule
Architecture & Decoder-only & Decoder-only & Decoder-only \\
LayerNorm position & Post-LN & Pre-LN & Pre-LN \\
Positional embeddings & Absolute & RoPE & RoPE \\
Attention mechanism & MHA & GQA/MQA & MHA \\
Residual connection & Sequential & Parallel & Parallel \\
Activation function & GeLU & SwiGLU & GeLU \\
\midrule
Model size & 124M & 1.3B & 2B \\
Context length & 1k & 16k & 8k \\
\midrule
Training objective & Next-token prediction & Next-token + multi-token & Knowledge distillation \\
Memory optimization & Standard & Speed + memory efficient & Knowledge distill. + memory \\
\bottomrule
\end{tabular}
\end{center}

\begin{center}
\parbox{\textwidth}{\captionof{table}{Architectural comparison of language models used in our experiments. All models follow the decoder-only transformer architecture with key differences in normalization, attention mechanisms, and optimization strategies.}}
\end{center}
\label{tab:model_comparison}


\clearpage
\begin{figure*}
    \centering
    \textbf{Adjective}\par\medskip
    \begin{subfigure}[b]{0.32\linewidth}
        \includegraphics[width=\linewidth]{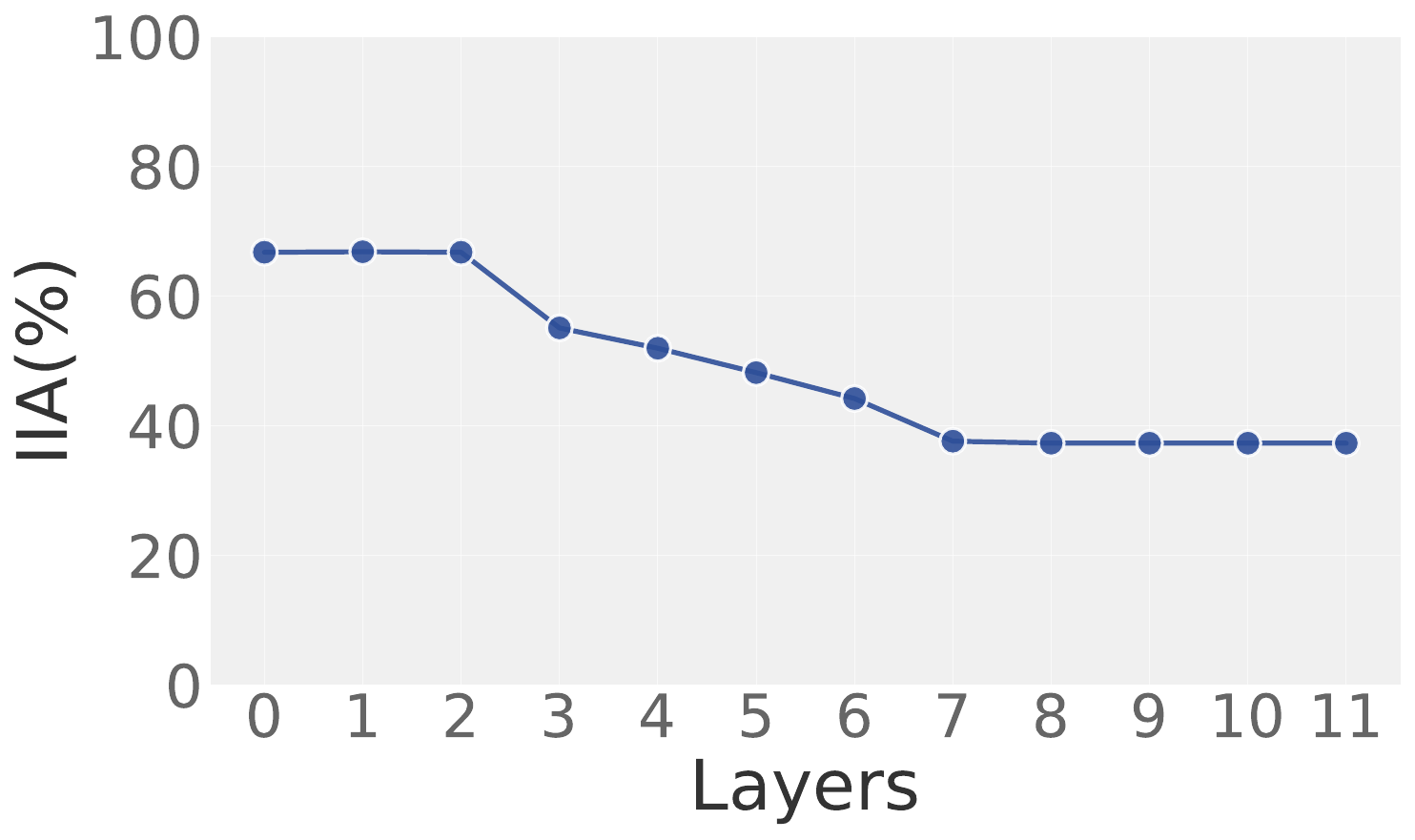}
        \caption{GPT-2}
    \end{subfigure}
    \begin{subfigure}[b]{0.32\linewidth}
        \includegraphics[width=\linewidth]{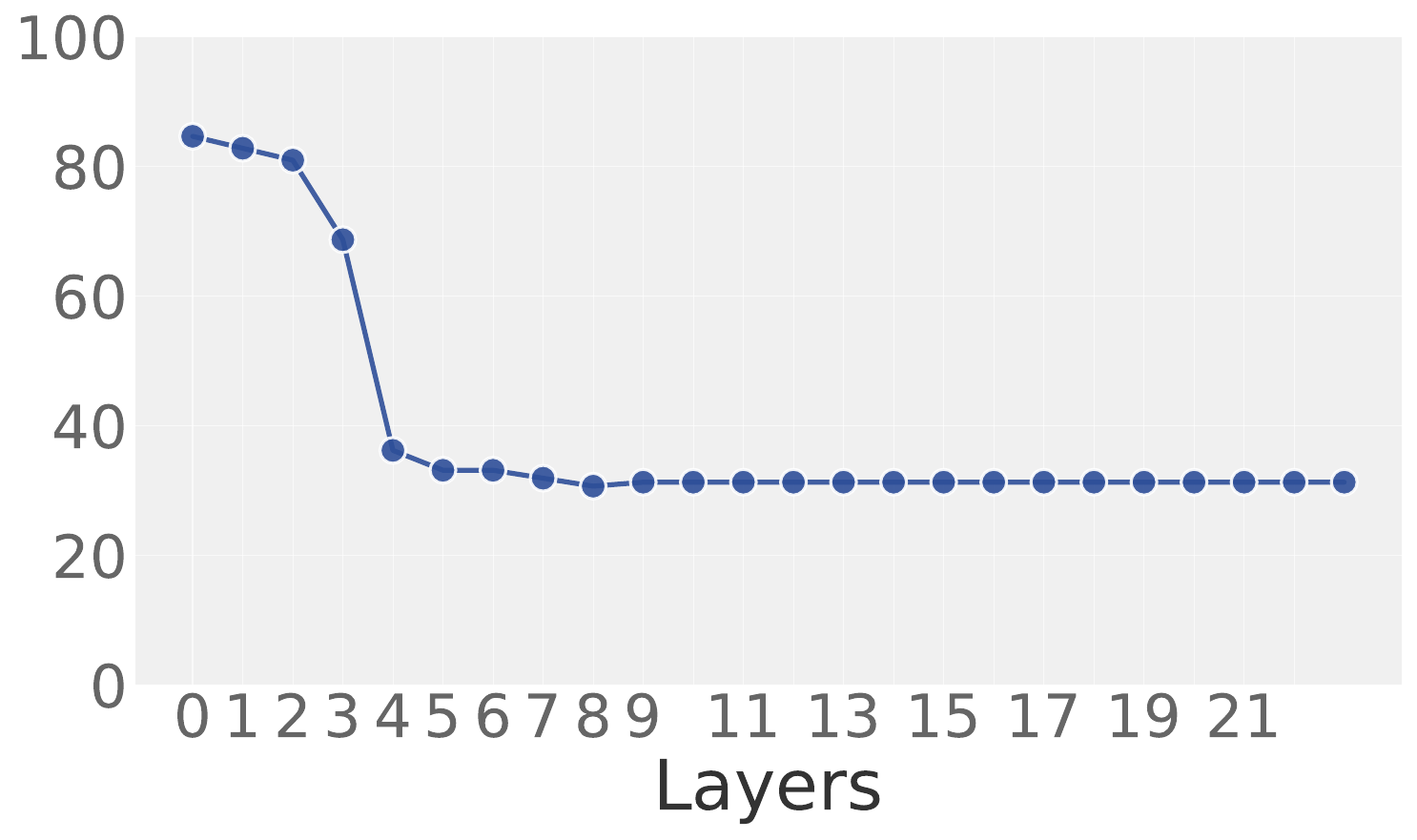}
        \caption{DeepSeek}
    \end{subfigure}
    \begin{subfigure}[b]{0.32\linewidth}
        \includegraphics[width=\linewidth]{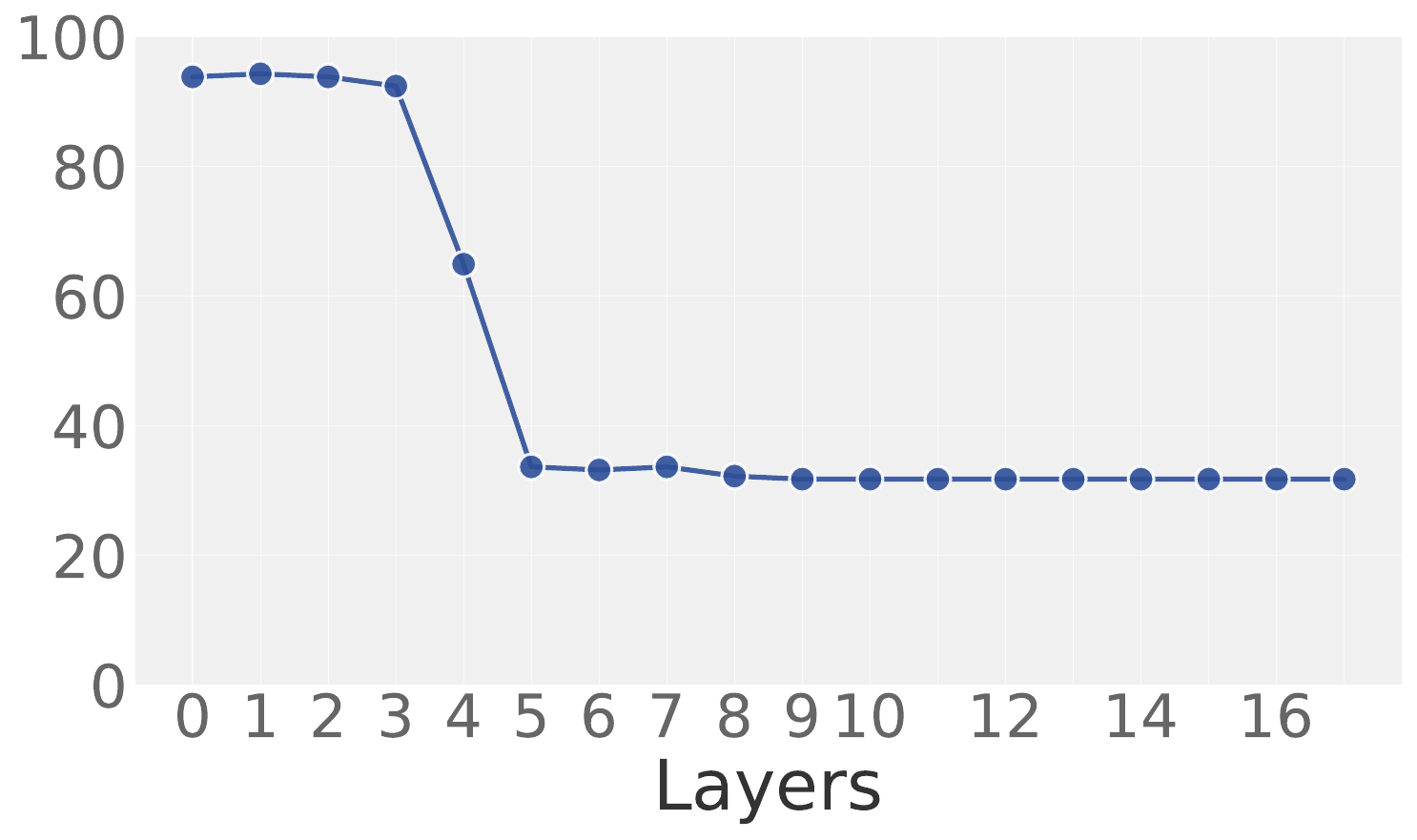}
        \caption{Gemma}
    \end{subfigure}

    \vspace{0.5cm}
    \begin{subfigure}[b]{\textwidth}
        \centering
        \textbf{Subject}\par\medskip
        \begin{subfigure}[b]{0.32\textwidth}
            \includegraphics[width=\linewidth]{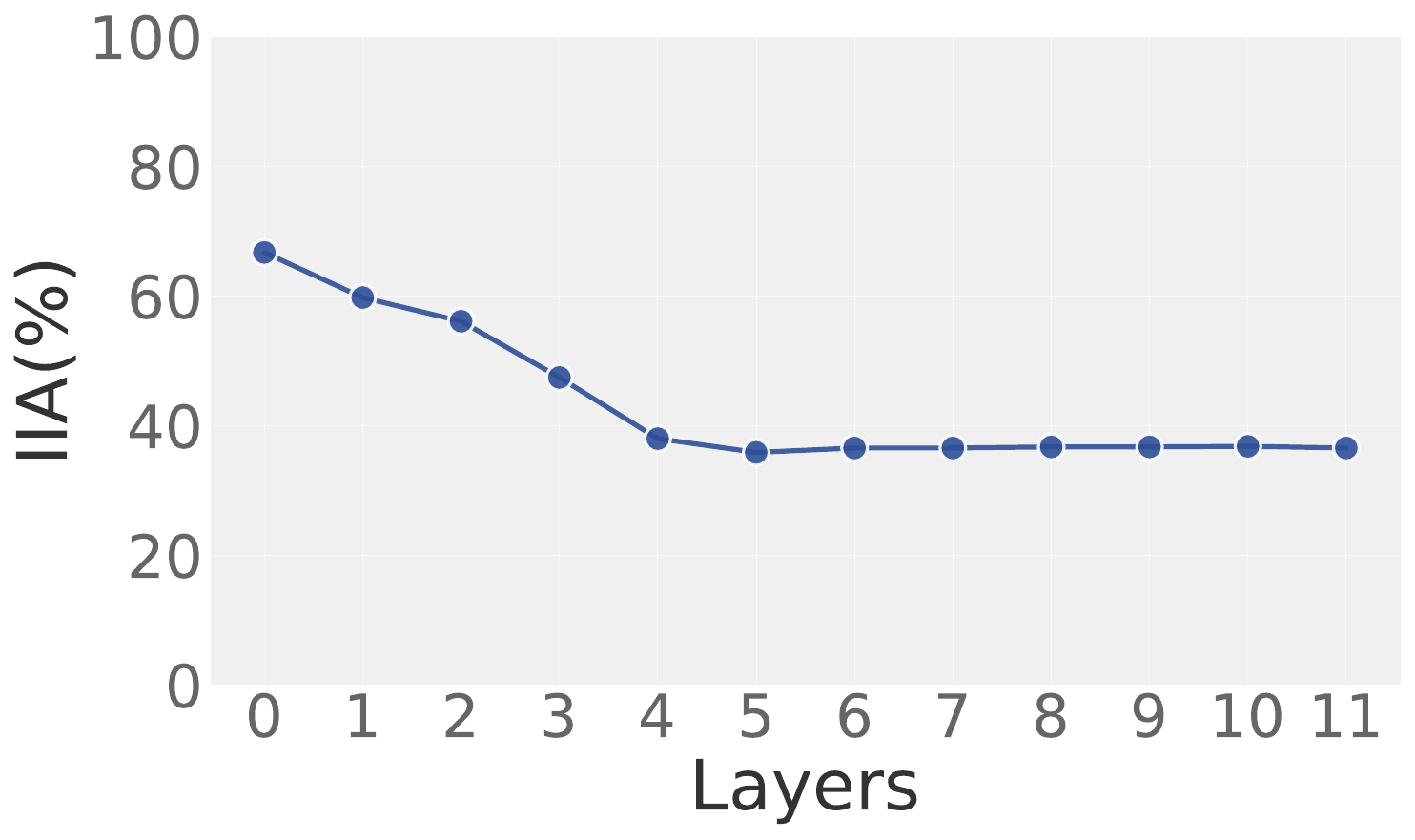}
            \caption{GPT-2}
        \end{subfigure}
        \begin{subfigure}[b]{0.32\textwidth}
            \includegraphics[width=\linewidth]{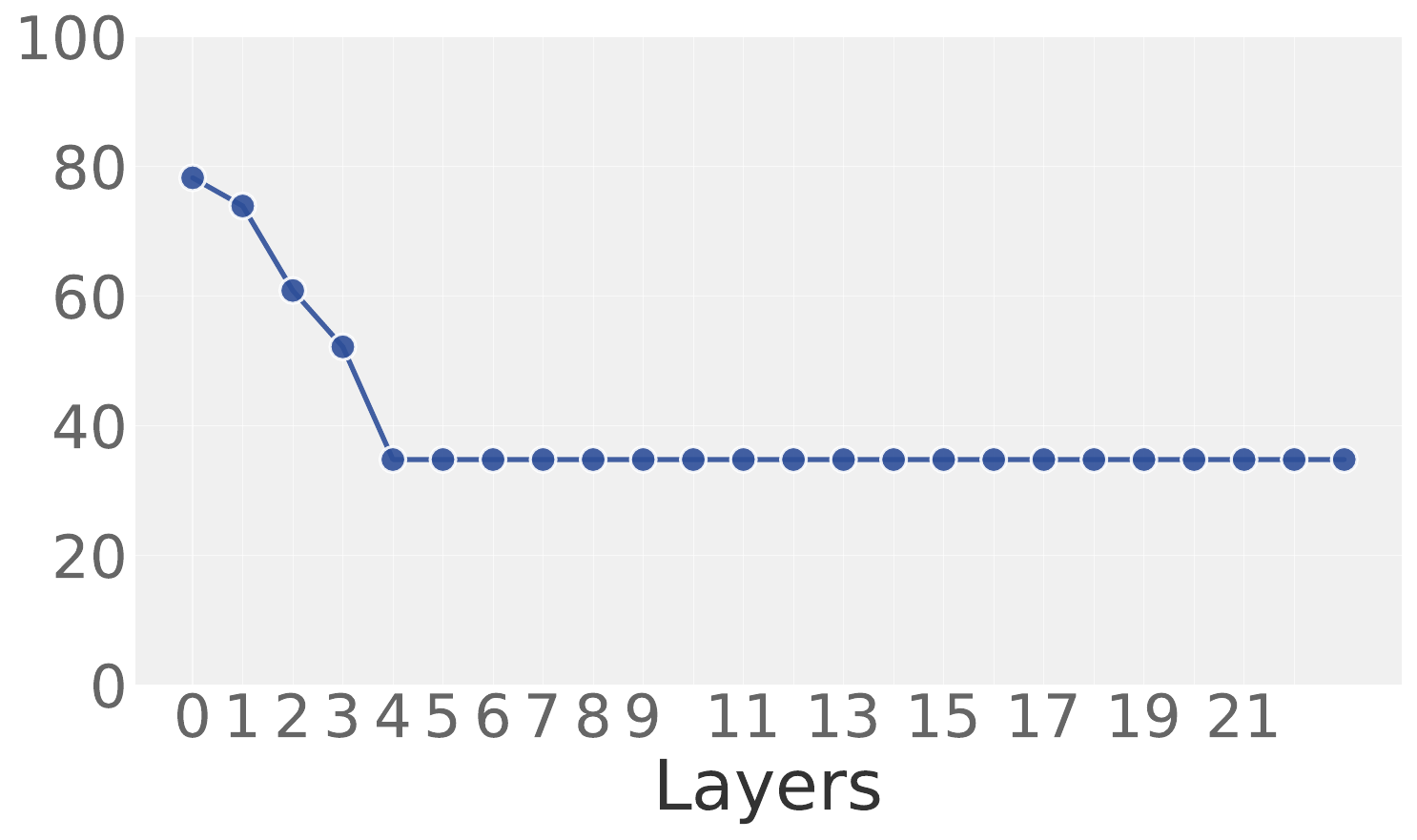}
            \caption{DeepSeek}
        \end{subfigure}
        \begin{subfigure}[b]{0.32\textwidth}
            \includegraphics[width=\linewidth]{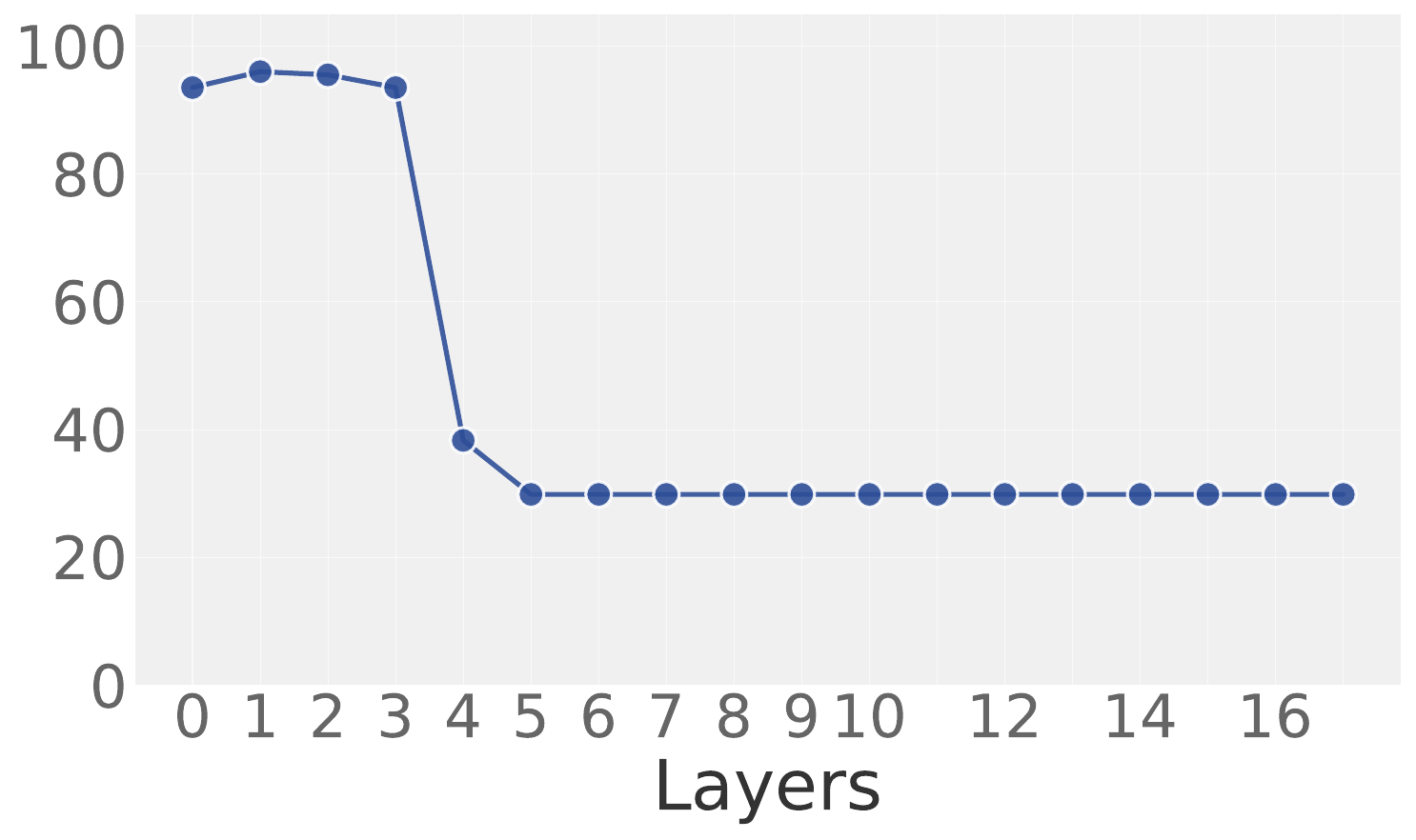}
            \caption{Gemma}
        \end{subfigure}
    \end{subfigure}
    
    \vspace{0.5cm}
    \begin{subfigure}[b]{\textwidth}
        \centering
        \textbf{Second Sentence}\par\medskip
        \begin{subfigure}[b]{0.32\textwidth}
            \includegraphics[width=\linewidth]{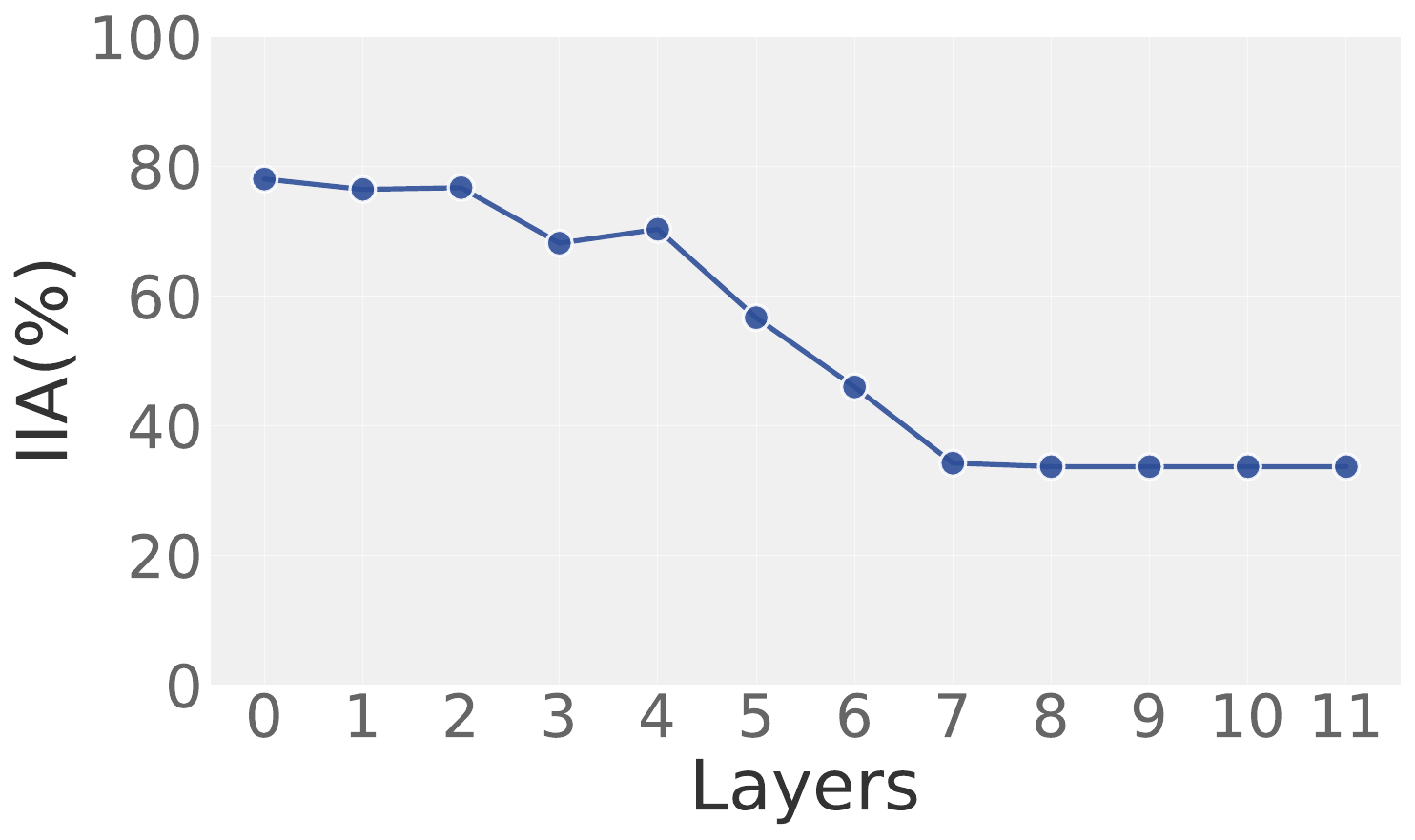}
            \caption{GPT-2}
        \end{subfigure}
        \begin{subfigure}[b]{0.32\textwidth}
            \includegraphics[width=\linewidth]{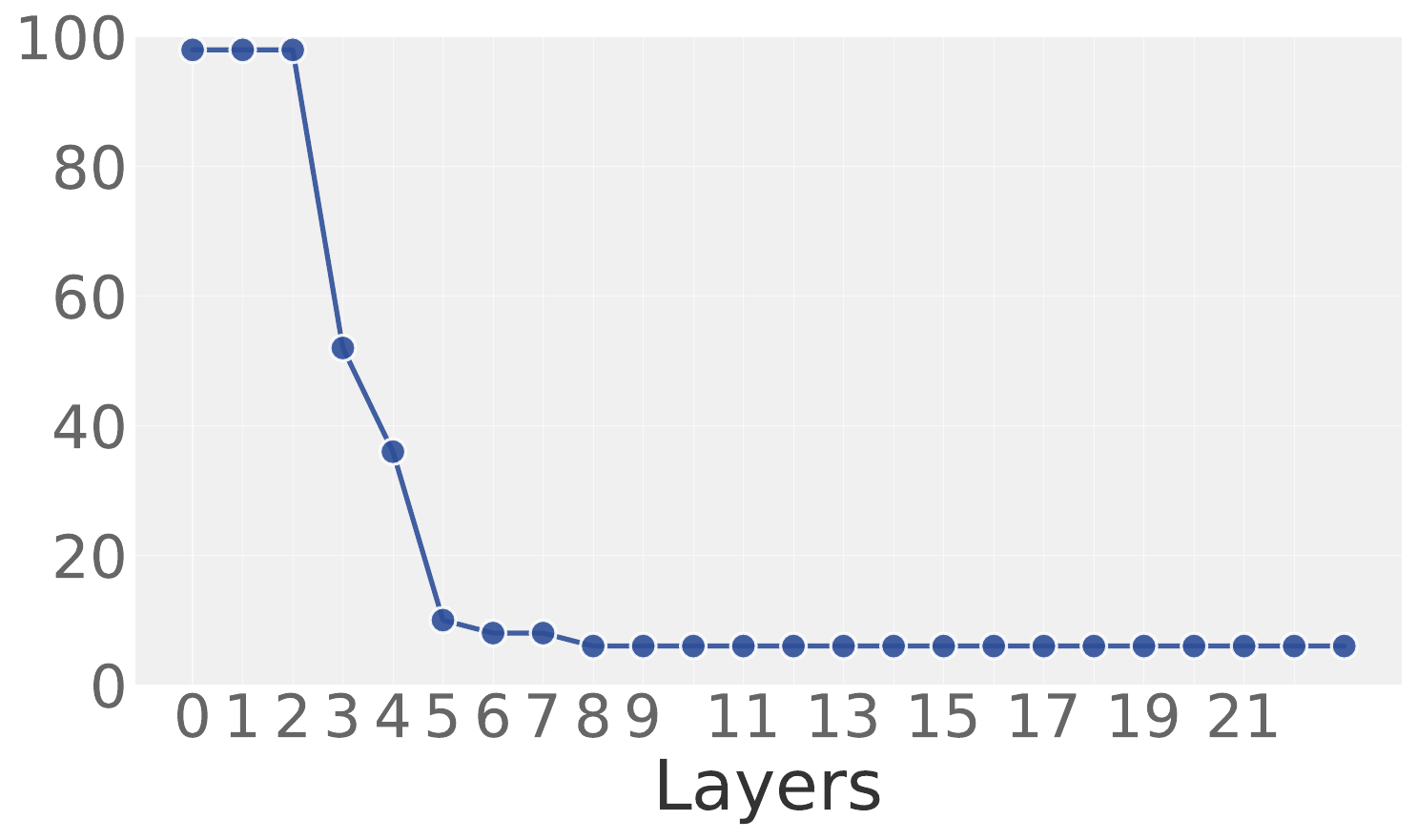}
            \caption{DeepSeek}
        \end{subfigure}
        \begin{subfigure}[b]{0.32\textwidth}
            \includegraphics[width=\linewidth]{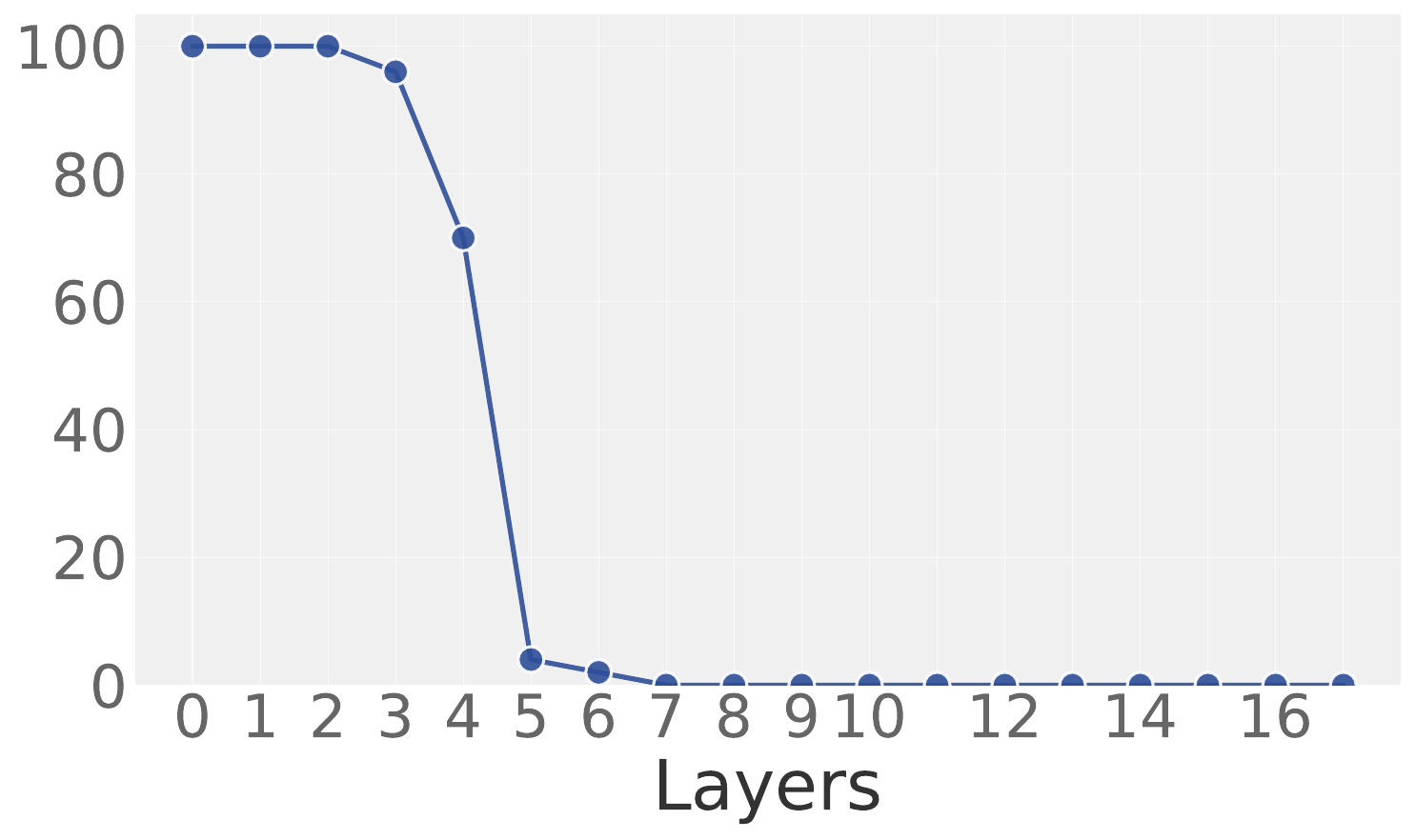}
            \caption{Gemma}
        \end{subfigure}
    \end{subfigure}
    
    \vspace{0.5cm}
    \begin{subfigure}[b]{\textwidth}
        \centering
        \textbf{Second Period}\par\medskip
        \begin{subfigure}[b]{0.32\textwidth}
            \includegraphics[width=\linewidth]{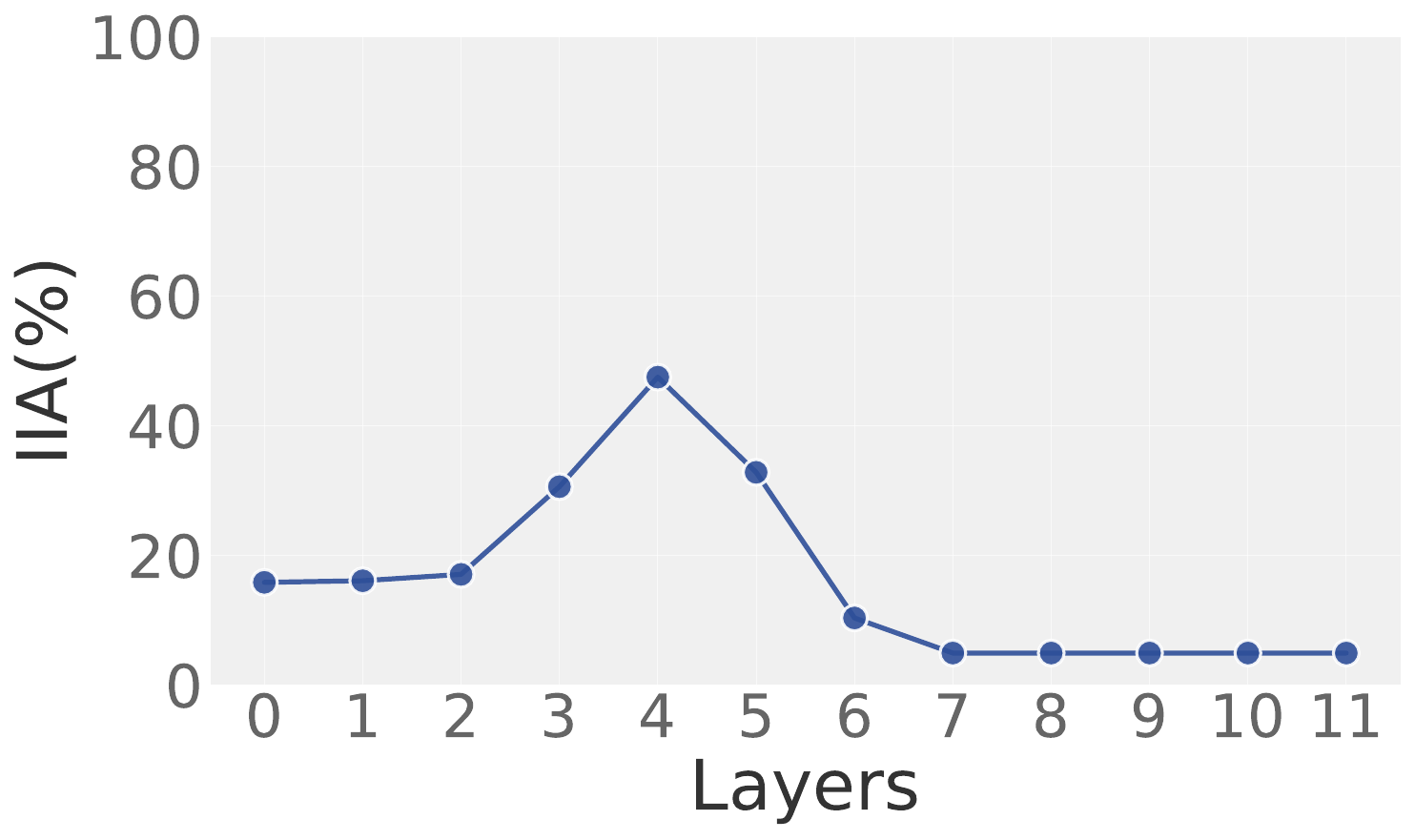}
            \caption{GPT-2}
        \end{subfigure}
        \begin{subfigure}[b]{0.32\textwidth}
            \includegraphics[width=\linewidth]{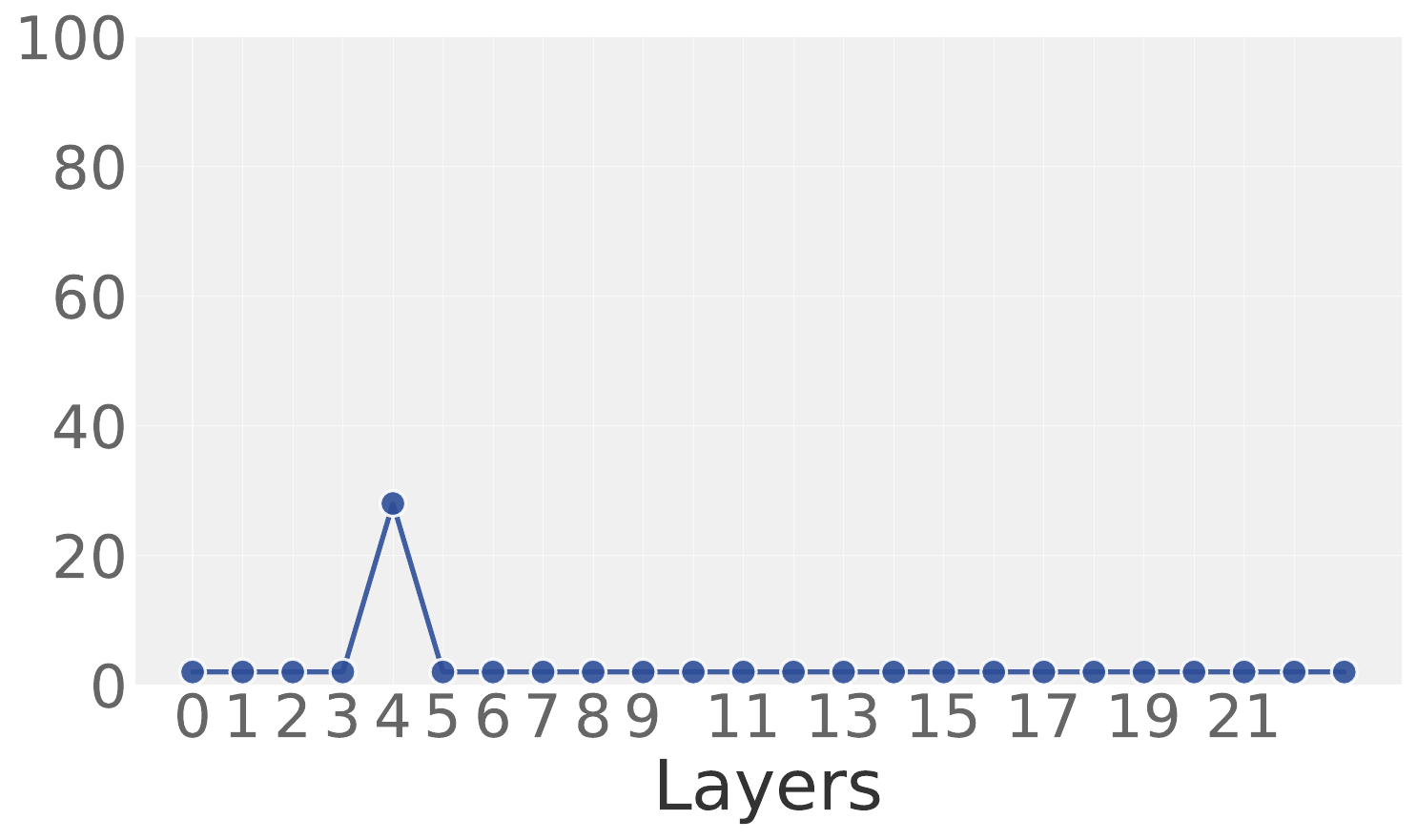}
            \caption{DeepSeek}
        \end{subfigure}
        \begin{subfigure}[b]{0.32\textwidth}
            \includegraphics[width=\linewidth]{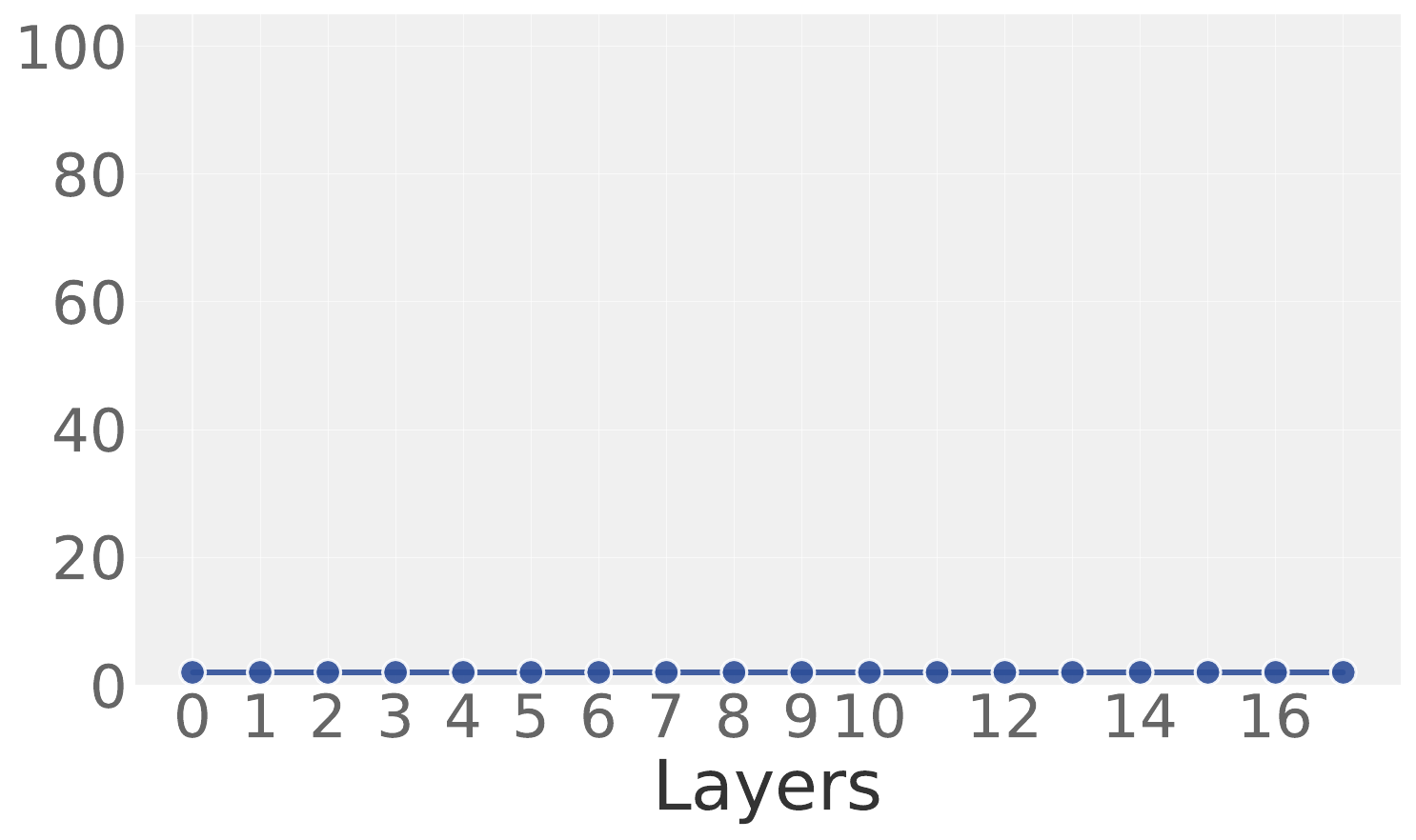}
            \caption{Gemma}
        \end{subfigure}
    \end{subfigure}

    \caption{Layer-wise sensitivity to logical rule and period token interchange interventions across GPT-2, DeepSeek, and Gemma. Each row represents a rule type; each column corresponds to a model. 
    }
    \label{fig:additional-IIA-results}
\end{figure*}

\end{document}